\documentclass{article}

\usepackage{arxiv}
\usepackage{authblk}
\usepackage{hyperref}       
\usepackage{url}            
\usepackage{booktabs}       
\usepackage{amsfonts}       
\usepackage{nicefrac}       
\usepackage{microtype}      
\usepackage{graphicx}
\usepackage[T1]{fontenc}
\usepackage[utf8]{inputenc}
\usepackage{amsmath}
\usepackage{rotating}
\usepackage{multirow}
\usepackage{cite}
\graphicspath{ {./figures/} }

\title{SAGE-Amine: Generative Amine Design with Multi-Property Optimization for Efficient CO$_{2}$ Capture}
\author[a]{Hocheol Lim$^{\ast}$}
\author[b]{Hyein Cho$^{}$}
\author[a]{Jeonghoon Kim$^{}$}

\affil[a]{Bioinformatics and Molecular Design Research Center (BMDRC), Incheon, 21983, Republic of Korea}
\affil[b]{The Interdisciplinary Graduate Program in Integrative Biotechnology \& Translational Medicine, Yonsei University, Incheon, 21983, Republic of Korea}

\begin{document}
\maketitle

\begin{flushleft}
    $^{\ast}$ Corresponding author: Hocheol Lim (ihc0213@yonsei.ac.kr)
\end{flushleft}

\vspace{0.5cm}

\begin{abstract}
Efficient CO\textsubscript{2} capture is vital for mitigating climate change, with amine-based solvents being widely used due to their strong reactivity with CO\textsubscript{2}. However, optimizing key properties such as basicity, viscosity, and absorption capacity remains challenging, as traditional methods rely on labor-intensive experimentation and predefined chemical databases, limiting the exploration of novel solutions. Here, SAGE-Amine was introduced, a generative modeling approach that integrates Scoring-Assisted Generative Exploration (SAGE) with quantitative structure-property relationship models to design new amines tailored for CO\textsubscript{2} capture. Unlike conventional virtual screening restricted to existing compounds, SAGE-Amine generates novel amines by leveraging autoregressive natural language processing models trained on amine datasets. SAGE-Amine identified known amines for CO\textsubscript{2} capture from scratch and successfully performed single-property optimization, increasing basicity or reducing viscosity or vapor pressure. Furthermore, it facilitated multi-property optimization, simultaneously achieving high basicity with low viscosity and vapor pressure. The 10 top-ranked amines were suggested using SAGE-Amine and their thermodynamic properties were further assessed using COSMO-RS simulations, confirming their potential for CO\textsubscript{2} capture. These results highlight the potential of generative modeling in accelerating the discovery of amine solvents and expanding the possibilities for industrial CO\textsubscript{2} capture applications.
\end{abstract}


\section{Introduction}
The rapid industrial development and increasing global energy demand have led to sustained reliance on fossil fuels, causing a significant rise in CO\textsubscript{2} emissions and disrupting the natural carbon cycle \cite{ref_1, ref_2}. These emissions are a major driver of the greenhouse effect and global warming, creating urgent environmental challenges. The natural capacity of oceans to absorb atmospheric CO\textsubscript{2} is insufficient to cope with these elevated emission levels \cite{ref_2}, making additional mitigation strategies essential. Carbon capture and storage technologies have gained prominence, with amine-based solvents standing out for post-combustion CO\textsubscript{2} capture due to their compatibility with existing industrial processes \cite{ref_3, ref_4}.

Amine-based solvents, such as monoethanolamine (MEA), diethanolamine (DEA), methyldiethanolamine (MDEA), and piperazine (PZ), have been extensively studied for post-combustion CO\textsubscript{2} capture \cite{ref_5, ref_6, ref_7, ref_8}. These solvents react chemically with CO\textsubscript{2}, forming reversible bonds that allow for efficient gas separation. Primary and secondary amines form carbamates upon reacting with CO\textsubscript{2}, whereas tertiary amines capture CO\textsubscript{2} as bicarbonate \cite{ref_4}. The formation of bicarbonate by tertiary amines requires a lower heat of reaction than carbamate formation, resulting in energy savings during regeneration \cite{ref_9}. However, these methods encounter economic and environmental challenges, including chemical degradation, corrosivity, toxicity, and high energy demands for solvent regeneration \cite{ref_10}. These drawbacks underscore the critical need for discovering new amines with enhanced properties to improve the efficiency and cost-effectiveness of CO\textsubscript{2} capture processes.

Recent research has aimed to improve these solvents by focusing on developing blends or novel amine formulations that offer a better balance between CO\textsubscript{2} absorption capacity, stability, and energy efficiency \cite{ref_3, ref_11}. While primary amines like MEA react quickly with CO\textsubscript{2}, tertiary amines such as MDEA provide better regeneration efficiency but lower reactivity, leading to the development of mixed amine systems to capitalize on the strengths of each class. Studies have also delved into the modification of amine structures through the introduction of sterically hindered amines, polyamines, or amine-functionalized materials that can reduce regeneration energy while maintaining high CO\textsubscript{2} absorption capacity \cite{ref_11}. Despite these advancements, the process of optimizing amine-based solvents remains labor-intensive, relying heavily on experimental trial and error.

To address these challenges, computational methods have emerged as a promising alternative for accelerating the discovery and optimization of new amine solvents. Efforts to identify the optimal amine for CO\textsubscript{2} capture, based on absorption and desorption potential, have predominantly utilized virtual screening from existing databases or enumeration of a limited set of known amines \cite{ref_4, ref_12, ref_13}. However, the advent of de novo molecular design through generative deep learning has revolutionized the creation of molecules from scratch, enabling the generation of molecules with specific and desired properties. This approach directly learns molecular structures and properties from input data, thereby eliminating the need for predefined rules. Significant advancements in natural language processing (NLP) and machine learning algorithms have been pivotal in this progress. This innovative method has shown remarkable success in exploring vast chemical spaces and designing novel molecules with tailored properties \cite{ref_14, ref_15, ref_16}. For instance, long-short-term memory (LSTM) networks and genetic algorithms (GA) have been used to generate new molecules, surpassing those in existing databases, as demonstrated by benchmarks like GuacaMol \cite{ref_17}. The scoring-assisted generative exploration (SAGE) method has been developed to design molecules with desired properties \cite{ref_18}. This is achieved by integrating LSTM and GA, using the bridged bicyclic ring and virtual synthesis operators, and incorporating various quantitative structure-activity/property relationship (QSAR/QSPR) models. These generative models facilitate efficient exploration of chemical spaces, leading to the discovery of novel molecules.

In this study, the SAGE framework was extended to specifically target the development of new amine solvents for CO\textsubscript{2} capture, referred to as SAGE-Amine. This approach integrates advanced generative modeling methods with QSPR models to generate and evaluate novel amines for high CO\textsubscript{2} absorption. Initially, the SAGE-Amine models were pre-trained on extensive datasets of various amines. Leveraging autoregressive NLP methods with LSTM, Transformer, vanilla Transformer Decoder (TD), and modified Transformer Decoder (X-Transformer Decoder, XTD), the quality of molecules generated by the pre-trained models were compared. Secondly, benchmarking with the pre-trained models was performed to identify known amines for CO\textsubscript{2} capture and to generate new amines with specific molecular formulas. Thirdly, SAGE-Amine's capabilities were assessed through single-property optimization (SPO) tasks, focusing on achieving high pKa, low viscosity, and low vapor pressure separately in primary and secondary amines, tertiary, cyclic and polyamines, and even irrespective of amine types. Additionally, multiple-property optimization (MPO) tasks were conducted to target high CO\textsubscript{2} absorption along with high pKa, low viscosity, low vapor pressure, high water solubility, high synthetic accessibility, low cost, and moderate boiling and melting points. Lastly, quantum-chemical simulations using COSMO-RS were performed, confirming the potential of the generated amines. The results demonstrated that SAGE-Amine could effectively navigate the chemical space of amines and optimize multiple properties simultaneously, making it a powerful tool for early-stage material discovery.

\section{Methods}
\subsection{Scoring-Assisted Generative Exploration for Amines (SAGE-Amine)}
Scoring-assisted generative exploration (SAGE) utilizes an iterative fine-tuning approach to create new molecules \cite{ref_18, ref_19}. This is achieved through autoregressive NLP and chemical diversification operators. The generated molecules are then evaluated using various scoring models to ensure they meet pre-defined objectives, thereby producing high-scoring molecules. The SAGE was extended for Amines (SAGE-Amine) by pre-training LSTM \cite{ref_20, ref_21}, Transformer \cite{ref_22}, vanilla Transformer Decoder (TD) \cite{ref_23}, and modified Transformer Decoder (X-Transformer Decoder, XTD) \cite{ref_24, ref_25, ref_26, ref_27, ref_28, ref_29, ref_30, ref_31, ref_32, ref_33, ref_34, ref_35, ref_36, ref_37, ref_38} models. Furthermore, amine-specific QSPR scoring models were integrated into SAGE-Amine.

\begin{table}[ht!]
    \centering
    \caption{Summary of Datasets used in this work}
    \renewcommand{\arraystretch}{1.3} 
    \begin{tabular}{@{}c c c c@{}}
        \toprule
        \textbf{Class} & \textbf{Task} & \textbf{Unit} & \textbf{All set} \\
        \specialrule{1.5pt}{0pt}{0pt}
        \multirow{4}{*}{Pre-train} & Amine-250 & \multirow{4}{*}{Count} & 706,279 \\
        & Amine250-reduced & & 701,114 \\
        & Amine300 & & 2,133,461 \\
        & Amine300-reduced & & 2,127,298 \\
        \cmidrule(lr){1-4}
        \multirow{4}{*}{QSPR} & Viscosity & log\textsubscript{10}(cP) & 3582 \\
        & Vapor Pressure & log\textsubscript{10}(mmHg) & 2945 \\
        & Boiling Point & ℃ & 23,044 \\
        & Melting Point & ℃ & 9721 \\
        \bottomrule
    \end{tabular}
    \label{tab:table_1}
\end{table}

The amines utilized in SAGE-Amine are encoded as character sequences in the Simplified Molecular Input Line Entry System (SMILES) format \cite{ref_39}. The amine structures used for pre-training the SAGE-Amine models were sourced from various QSPR datasets \cite{ref_40, ref_41, ref_42, ref_43, ref_44} and in-stock ZINC20 \cite{ref_45}, as summarized in Table~\ref{tab:table_1}. The amines were categorized into two groups by molecular weight: those with a weight of 250 or less (Amine250) and those with a weight of 300 or less (Amine300). To avoid similarity with target amines used in goal-directed benchmarks, amines with a maximum similarity exceeding 0.323 were excluded, as calculated by the Morgan fingerprint in RDKit \cite{ref_46}. This exclusion resulted in reduced datasets (Amine250-reduced and Amine300-reduced). These four datasets were then randomly split into training and validation sets, with proportions of 0.998 and 0.002, respectively.

The NLP components in SAGE-Amine for autoregressive generation include LSTM, Transformer, TD, and XTD models. The LSTM model comprises three layers with 1,024 hidden units, a dropout probability of 0.2, a learning rate of 0.001, and a batch size of 512. The Transformer model was trained with eight attention heads, three encoder and decoder layers, 1,024 hidden units, a dropout probability of 0.2, an embedding size of 256, a learning rate of 0.001, and a batch size of 512. The TD and XTD models were trained similarly but utilized only the decoder layers.

The XTD incorporates several modifications to the TD model \cite{ref_24, ref_25, ref_26, ref_27, ref_28, ref_29, ref_30, ref_31, ref_32, ref_33, ref_34, ref_35, ref_36, ref_37, ref_38}. Firstly, it includes 20 learned memory tokens that are processed through the attention layers alongside the input tokens to alleviate outliers \cite{ref_25}. Secondly, it employs layer normalization through L2-normalized embeddings to improve convergence \cite{ref_26}. Within the attention layer, it integrates rotary embeddings and T5 simplified embeddings with relative positional embeddings \cite{ref_27, ref_28}. It shifts a subset of the feature dimension along the sequence dimension by one token to aid convergence \cite{ref_30}. Furthermore, the XTD gates the residual connections within the transformer network to enhance stability and performance \cite{ref_31}. It utilizes a single head for keys and values but maintains multi-headed queries for memory efficiency \cite{ref_33}. The queries and keys are L2-normalized along the head dimension before the dot product (cosine similarity) to prevent the attention operation from overflowing \cite{ref_34}. An efficient method is used to sparsify attention by zeroing all dot-product query/key values that do not fall within the top k values \cite{ref_35}. Information is mixed between heads both before and after attention (softmax) \cite{ref_36}. The normformer is applied to provide per-head scaling after aggregating the values in attention, which aids convergence, and an extra layer normalization is applied right after the activation in the feedforward process for performance improvement \cite{ref_32}. Gated linear unit variants are included for performance enhancements \cite{ref_37}, and there is no bias in the feedforward layers to increase throughput without any loss of accuracy \cite{ref_38}.

The NLP models were pre-trained with the Adam optimizer \cite{ref_47} over 150 epochs across four datasets. The final model weights were selected based on achieving the lowest loss on the validation sets. The generated amines were assessed using several performance metrics: validity, uniqueness, novelty, diversity, and the ratios of amines and amine types (primary, secondary, tertiary, cyclic, and polyamines). Validity evaluates the model’s capability to integrate chemically accurate constraints and syntax in SMILES, ensuring correct valence. Uniqueness measures the model's ability to produce a varied set of amines, avoiding a limited range. Novelty is gauged by the percentage of generated amines absent from the training data. Diversity is evaluated by analyzing the chemical variation within the generated molecule sets, determined using sphere exclusion diversity (SEDiv \cite{ref_48}). The ratios of amines refer to the proportion of molecules containing an amine group within the generated set. The ratio of amine types reflects the distribution of these amines into their respective categories based on the number of hydrogen atoms attached to the nitrogen atom. Notably, even in cases of polyamines, the presence of an amine within a ring structure classifies it as cyclic.

The pre-trained NLP models in SAGE-Amine generate amines autoregressively in each iteration, first verifying the generated amines to ensure they have valid SMILES strings and do not contain radicals. Next, the molecules are checked to confirm if they are amines and, if so, to determine their specific amine type. Following the SAGE framework \cite{ref_18}, chemical diversification operators, including mutation and crossover, are then applied. The mutation operator introduces various chemical modifications at the atomic level, such as appending atoms, inserting atoms, altering bond orders in covalent bonds, adding or removing ring bonds, and creating bridged bicyclic rings in a ring substructure. The crossover operator randomly splits a pair of parent molecules into fragments and recombines them to form a new molecule at the functional group level, even allowing attachment to bridgehead atoms in bridged bicyclic rings.

After chemical diversification, the molecules are re-evaluated to confirm their status as amines. These amines are then assessed according to predefined objectives, are ranked based on their scores, and the top 1,024 amines are stored in a buffer. This buffer is used to fine-tune the NLP model with a learning rate of 0.001 and a batch size of 256 for 8 epochs. The GA and NLP-only models (LSTM, TD, and XTD) each generate 16,384 molecules. In the LSTM/GA, TD/GA, and XTD/GA models, the NLP model generates 8,192 molecules while the GA produces an additional 8,192. The GA-only models initially select 16,384 amines randomly from the training data and generate 16,384 molecules. In all tasks, SAGE-Amine generates molecules over 100 iterations.

\subsection{Goal-Directed Benchmarks and Score Definition in SAGE-Amine} 
Goal-directed benchmarks for evaluating the performance of generative models were established following the protocols used in the GuacaMol \cite{ref_17} and SAGE \cite{ref_18} studies. Rediscovery tasks involved identifying specific target amines, namely 2-aminoethanol (MEA), 2-amino-2-hydroxymethyl-1,3-propanediol (AHMPD), diethanolamine (DEA), diethylamine (DA), N,N-diethylethanolamine (DEEA), N-methyldiethanolamine (MDEA), 2-methylpiperazine (2-MPZ), 2-piperidineethanol (2-PPE), homopiperazine (HomoPZ), and diethylenetriamine (DETA). Similarity tasks required generating amines similar to a target amine, selecting the top 100 generated amines with a similarity score above 0.7, and using their average similarity for evaluation. The amines used for similarity tasks included 2-amino-2-methyl-1-propanol (AMP), isopropylamine (IPA), 3-(isopropylamino)propanol (IPAP), 4-dimethylamino-1-butanol (4DMA1B), 1-dimethylamino-2-propanol (1DMA2P), 1-methyl-piperazine (1-MPZ), 1-ethyl-piperazine (EPZ), piperazine (PZ), 1-(2-aminoethyl)piperazine (AEP), and Triethylenetetramine (TETA). Median similarity tasks used the average similarity value for two target amines, again selecting the top 100 with a similarity score above 0.7. The amines used for these tasks were 2-(2-diethylaminoethoxy)ethanol (DEAE-EO), 1-methyl-2-piperidinemethanol (1M-2PPE), and 1-(2-hydroxyethyl)piperdine (1-(2HE)PP). Isomer tasks involved generating ions that matched a specified molecular formula, addressing the issue of overfitting by ensuring the production of diverse ions rather than those following a simple pattern. The molecular formulas of C\textsubscript{4}H\textsubscript{11}NO originated from AMP, C\textsubscript{4}H\textsubscript{11}NO\textsubscript{2} from IPAP, DEEA, and 4DMA1B, and C\textsubscript{5}H\textsubscript{12}N\textsubscript{2} from 1-MPZ, 2-MPZ, and HomoPZ. For all goal-directed benchmarks, similarity scores were computed using the Extended-connectivity fingerprints (ECFP6) \cite{ref_49}.

Single-property optimization (SPO) tasks involve the maximization of basic pKa and the minimization of viscosity or vapor pressure in amines by generating amines using SAGE-Amine. Additionally, to perform SPO tasks based on amine type, three separate generations were conducted: one for primary and secondary amines, one for tertiary, cyclic, and polyamines, and one without constraints. For cases where the generated amine did not match the desired type, a penalty was applied by reflecting only 10$\%$ of the score. The pKa values of all titratable sites were predicted within the molecule through MolGpKa \cite{ref_40}, where the average pKa value of the amine groups was used as the score. Viscosity and vapor pressure were predicted at a standard temperature of 298.15 K.

Multiple-property optimization (MPO) tasks were designed to maximize the CO\textsubscript{2} absorption of amines. To achieve this, a range of factors were considered including basicity (pKa), viscosity, vapor pressure, boiling point, melting point, aqueous solubility, synthetic accessibility, and chemical price. Similar to the SPO tasks, penalties were applied based on the type of amine, used average scores for pKa, and predicted viscosity and vapor pressure at 298.15 K. Next, several adjustments were performed to create weights for these various properties, determining maximum and minimum limits and evaluating them linearly within these ranges. For pKa, a value greater than 14 resulted in a score of 1, while a value less than 7 resulted in a score of 0. For viscosity, a value greater than 2 resulted in a score of 0, while a value less than -1 resulted in a score of 1. For vapor pressure, values above 3 scored 0, and values below -3 scored 1. For boiling point, values below 80 scored 0, and values above 250 scored 1. For melting point, values above 80 scored 0, and values below 40 scored 1. Aqueous solubility (log\textsubscript{10}mol/L) was predicted using SolTranNet \cite{ref_50}; predicted values (LogS, mol/L) below -4 scored 0, and values above 2 scored 1. To determine synthetic accessibility, the retrosynthetic accessibility score (RAscore) was employed \cite{ref_51}, a metric that rapidly estimates a molecule’s synthetic feasibility. A score closer to 1 indicates a higher probability of finding feasible retrosynthetic pathways, reflecting the molecule’s ease of synthesis. The chemical price was predicted through CoPriNet \cite{ref_52}; predicted values ($\$$/gram) below 0 were assigned a score of 1, while those above 10 received a score of 0. The MPO score was calculated as the average of these eight scores (pKa, viscosity, vapor pressure, boiling point, melting point, aqueous solubility, synthetic accessibility, and price), along with the target CO\textsubscript{2} absorption score.

\subsection{Quantitative Structure-Property Relationship (QSPR)}
The datasets for viscosity (log\textsubscript{10}cP), vapor pressure (log\textsubscript{10}mmHg), boiling point (℃), and melting point (℃) are summarized in Table~\ref{tab:table_1}. The viscosity data were sourced from Chew \textit{et al.} \cite{ref_41}, while the vapor pressure, boiling point, and melting point data were obtained from EPI Suite 4.11 \cite{ref_43}. Additionally, some boiling point data were manually supplemented from other sources \cite{ref_42, ref_44, ref_53}. The maximum and minimum values for each dataset are as follows: viscosity ranges from -1 to 1.4236, with a temperature spanning 227.45 to 404.10 K. Vapor pressure ranges from -20.7399 to 7, with a temperature spanning 205.15 to 448.15 K. Boiling point values range from -268.935 to 5900 ℃, and melting point values range from -219.61 to 3410 ℃.

Molecular fingerprints were employed to generate numerical features of chemicals. Based on PyFingerprint \cite{ref_54}, the 11 types of molecular fingerprints were created: Avalon, ECFP6, Extended, FCFP4, MACCS, Morgan, PCFP, rDesc, rPair, rTorsion, and Standard. The Avalon fingerprint utilizes a generator that enumerates specific paths and feature classes of the molecular graph \cite{ref_55}. Extended-connectivity fingerprints with a diameter of 6 (ECFP6) are designed for structure-activity relationship modeling, representing circular atom neighborhoods as 1024-bit keys \cite{ref_49}. The Extended fingerprint is a 1024-bit hashed fingerprint that considers rings and atomic properties. Function-class fingerprints with a diameter of 4 (FCFP4) are similar to ECFP and index pharmacophore-like roles of distinct atoms within molecules as 1024-bit keys. The Molecular Access System (MACCS) is a widely used fingerprint for assessing structural similarity using 166-bit MACCS keys \cite{ref_56}, while the PubChem system employs substructural fingerprints (PCFP) with 881-bit structural keys to represent chemical structures, facilitating similarity and neighbor searches \cite{ref_53}. The rDesc, rPair, and rTorsion descriptors are derived from RDKit \cite{ref_46} and include rd-descriptor, topological torsion, and atom-pair \cite{ref_57}. The Standard fingerprint is a 1024-bit hashed fingerprint that considers paths of a given length. Additionally, six fingerprints were concatenated with MACCS fingerprints: Avalon, ECFP6, Extended, FCFP4, PCFP, and Standard to create a comprehensive fingerprint set.

To develop QSPR models using these fingerprints, several regression algorithms were applied: Random Forest (RF), Light Gradient Boosting Machine (LGBM), and Extreme Gradient Boosting (XGB). These algorithms utilize decision trees to minimize overfitting and reduce variance. Each decision tree evaluates numerical features to produce continuous outputs, constructed sequentially and refined based on previous errors. To identify the most effective QSPR models, a grid search was conducted, considering three hyperparameters for LGBM, two for RF, four for XGB, and two for GB. Detailed hyperparameter tuning for the QSPR models is provided in Table~\ref{tab:table_s2}. For LGBM, the hyperparameters were booster type (boosting$\_$type), the number of gradient-boosted trees (n$\_$estimators), and the learning rate (learning$\_$rate) \cite{ref_58}. Those of RF were the number of trees (n$\_$estimators) and the maximum tree depth (max$\_$depth) \cite{ref_59}. Those of XGB included booster type (booster), the number of trees (n$\_$estimators), maximum tree depth (max$\_$depth), and learning rate (learning$\_$rate) \cite{ref_60}. A fixed random seed was used to perform a stratified split for 5-fold cross-validation, ensuring a balanced distribution by dividing the y-values into quintiles.

Additionally, a molecular graph was utilized through the application of a directed message-passing neural network (D-MPNN) to develop QSPR models. The molecular graph captures detailed atomic features such as atom symbol, degree, valence, formal charge, chirality, radical presence, hybridization, the number of hydrogen atoms, and aromaticity. It also captures bond characteristics including bond type, conjugation, ring presence, and stereo configuration. Atom nodes are represented by several one-hot encoded vectors capturing chemical characteristics: atom symbols (44-dimensional), degree (12-dimensional), valence (8-dimensional), formal charge (1-dimensional), chirality (5-dimensional), radical presence (1-dimensional), hybridization (6-dimensional), the number of hydrogen atoms (6-dimensional), and aromaticity (1-dimensional). Bond edges are similarly represented by vectors indicating bond type (5-dimensional for single, double, triple, aromatic, and unknown), conjugation (1-dimensional), ring presence (1-dimensional), and stereo configuration (7-dimensional). The D-MPNN updates node attributes by aggregating information from neighboring nodes; however, D-MPNN treats edges as directed \cite{ref_61}. It outputs the target property from the embedding for a molecule's sub-structure, highlighting essential features such as atom symbol, connectivity, and bond type. The hyperparameters for D-MPNN are the hidden units for the embedding (hidden$\_$unit) and the number of message-passing and updating rounds (steps).

\subsection{Conductor-like Screening Model for Real Solvents (COSMO-RS)}
The Conductor-like Screening Model for Real Solvents (COSMO-RS) is a quantum chemistry-based statistical thermodynamics model that enables the prediction of thermodynamic properties of fluids and liquid mixtures \cite{ref_62}. In COSMO-RS, the free energy of each species in solution is determined by calculating its chemical potential through a statistical thermodynamics algorithm, which iteratively evaluates system affinity to molecular surface polarity \cite{ref_63, ref_64}. Amines used in the goal-directed benchmark sets and the top-performing amines generated through SAGE-Amine were selected using COSMOtherm (COSMOlogic GmbH, Leverkusen, Germany). The structures of the amines were prepared using LigPrep in the Schrödinger suite \cite{ref_65}, and single conformers with the lowest ground-state energy were chosen for each amine. COSMO-RS calculations at the BP-TZVPD-FINE level were subsequently performed using TURBOMOLE \cite{ref_66}. These calculations involved full geometry optimization using density functional theory (DFT) with the Becke–Perdew (BP86) functional \cite{ref_67, ref_68} and DFT-D3 dispersion correction \cite{ref_69}, employing a grid size of m3 and the resolution of identity (RI) approximation \cite{ref_70}. The triple-$\zeta$ valence polarized (def-TZVP) basis set was used for the optimization, followed by single-point energy calculations with the same DFT/BP86-D3/RI method incorporating diffuse functions via the def2-TZVPD basis set. Thermodynamic analyses were conducted to determine CO\textsubscript{2} Henry’s law constant (MPa), CO\textsubscript{2} solubility (log\textsubscript{10}X\textsubscript{sol}), flash point (℃), vaporization enthalpy (kJ/mol), excess enthalpy (kJ/mol), and excess Gibbs free energy (kJ/mol) for a water-amine mixture (80wt$\%$ water, 20wt$\%$ amine) as well as the flash point of the pure amine. All calculations were performed at 323.15 K using the COSMOtherm modules.

\section{Results}

\subsection{Scoring-Assisted Generative Exploration for Amine (SAGE-Amine)}
Unlike traditional virtual screening, which identifies the best chemicals from known existing databases, generative models aim to design molecules from scratch that possess desired properties. Scoring-Assisted Generative Exploration (SAGE) alternates between generation and evaluation phases to discover new chemicals \cite{ref_18, ref_19}. During generation, molecules are created through autoregressive modeling and chemical diversification, while the evaluation phase iteratively refines the NLP model using multiple scoring metrics to optimize the generated compounds.

\begin{figure}[ht!]
  \centering
  \includegraphics[width=1.0\textwidth]{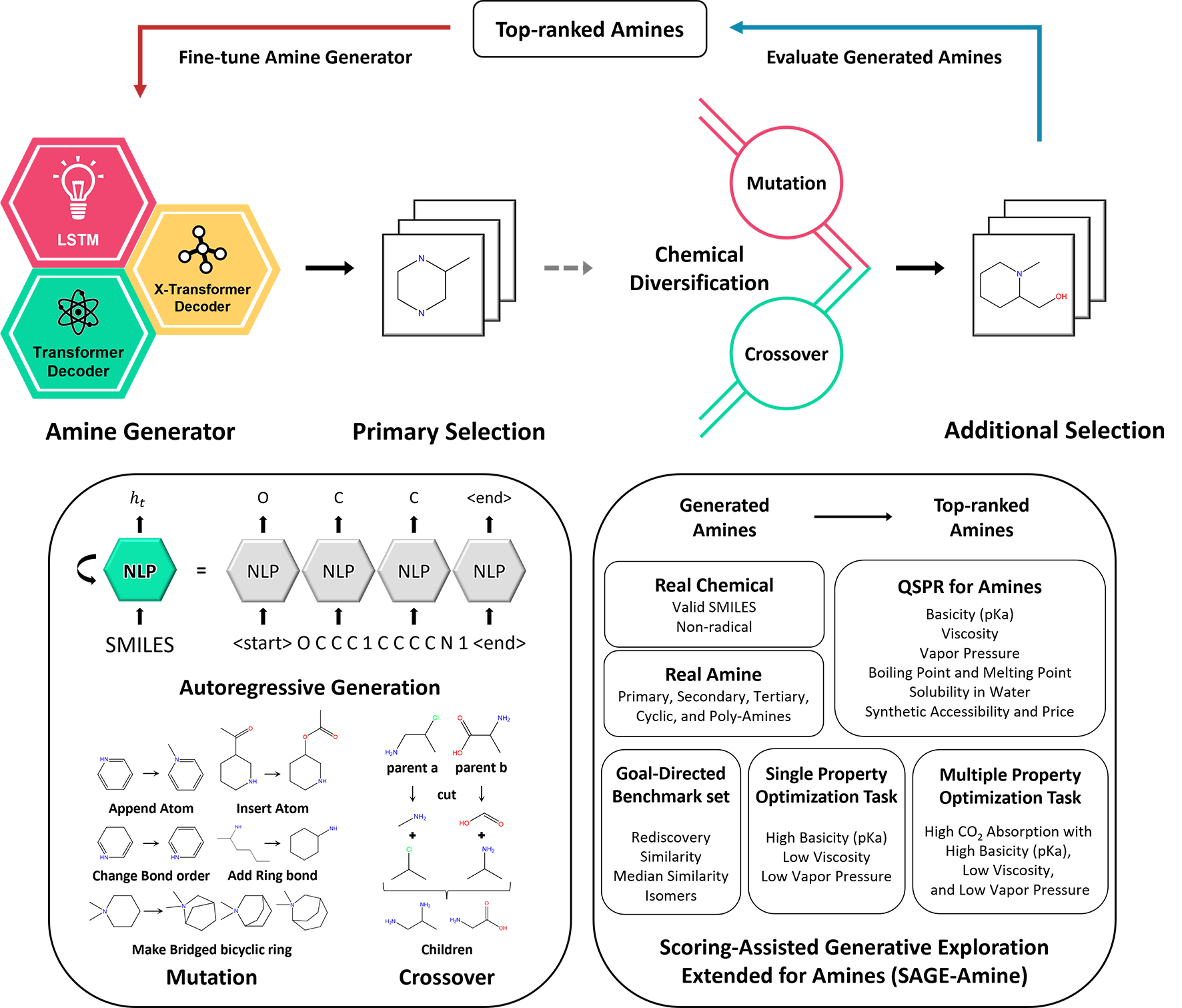}
  \caption{Scoring-Assisted Generative Exploration Extended for Amines (SAGE-Amine)}
  \label{fig:fig_1}
\end{figure}

For amine design in CO\textsubscript{2} absorption, this framework was extended into SAGE-Amine (Figure~\ref{fig:fig_1}), which operates with a generation and evaluation phase. The generation phase uses NLP models such as LSTM, Transformer, TD, and XTD, pre-trained on amine-specific datasets. These models generate token sequences autoregressively, creating new chemicals similar to compounds in pre-training datasets. Similar to the original SAGE, the generated molecules undergo further modifications via chemical diversification, broadening the chemical space.

In the evaluation phase, the generated molecules are validated to ensure they represent real chemicals by confirming valid SMILES, non-radically, and the presence of amine groups. Various properties relevant to efficient CO\textsubscript{2} absorption, such as basicity (pKa), viscosity, vapor pressure, boiling and melting points, aqueous solubility, synthetic accessibility, and cost, are predicted using QSPR models. The generated amines are ranked, and the top-ranked amines are selected to fine-tune the NLP models, enabling scoring-assisted generative exploration. This process facilitates both single-property optimization (SPO) and multiple-property optimization (MPO) tasks.

\subsection{Pre-training the SAGE-Amine with Amine databases}
To generate amines, SAGE-Amine required the NLP models, which learned the SMILES format and amine patterns through pre-training. Diverse amine data were gathered from sources like ZINC20 \cite{ref_45} and several QSPR datasets \cite{ref_40, ref_41, ref_42, ref_43, ref_44, ref_53}. The data was organized into four datasets based on molecular weight thresholds (Amine250, Amine250-reduced, Amine300, and Amine300-reduced), which are summarized in Table~\ref{tab:table_1}. Amine250 and Amine300 contained molecules with molecular weights below 250 and 300 g/mol, respectively, while the reduced datasets were created by filtering compounds based on their similarity to target amines in a goal-directed benchmark. The LSTM, Transformer, TD, and XTD algorithms were used to pre-train the models on these datasets.

Once pre-trained, these models generated 5,000 to 10,000 molecules, which were evaluated for validity, uniqueness, novelty, sphere-exclusion diversity, and whether they were amines (Table~\ref{tab:table_s1}). When trained on the Amine250 and Amine250-reduced datasets, all four models generated over 90$\%$ valid molecules. The LSTM and TD models showed high uniqueness, both producing over 95$\%$ unique molecules, while Transformer and XTD had lower uniqueness rates of 61$\%$ and 77$\%$, respectively. For novelty, the TD and XTD generated many new molecules not found in the training set, while the LSTM and Transformer produced more familiar compounds. The diversity of the generated molecules also varied. LSTM and TD excelled in diversity, while Transformer and XTD produced less diverse molecules. In terms of generating amines, LSTM, Transformer, and TD generated over 95$\%$ amines, while the XTD generated about 80$\%$. Only the XTD effectively generated both cyclic and more complex amines, while the other models mainly produced cyclic amines. When using the Amine300 and Amine300-reduced datasets, the models performed similarly, generating over 90$\%$ valid SMILES, although the XTD on Amine300-reduced had a lower validity rate of 50$\%$. The patterns of uniqueness, novelty, and diversity remained consistent, with LSTM, TD, and XTD generating diverse molecules.

From an algorithmic perspective, all models demonstrated the capability to generate valid amines, with distinct differences in novelty and diversity. The LSTM primarily produced cyclic amines from the training set, showing relatively low novelty. In contrast, the TD generated more novel molecules while still favoring cyclic amines. The XTD excelled, producing a diverse mix of cyclic and polyamines, along with high novelty. However, the Transformer model had low novelty, generating mainly known amines from the training set. From a dataset perspective, pre-training with the Amine300 dataset yielded better results compared to Amine250, with improved validity, uniqueness, and diversity. Among the reduced datasets, Amine250-reduced outperformed Amine300-reduced, offering a better balance between dataset size and optimization efficiency. Overall, the LSTM, TD, and XTD were identified as the most effective algorithms, with the Amine250-reduced and Amine300 datasets proving most suitable for pre-training. These results indicate that the combination of algorithm and dataset selection plays a critical role in generating novel, diverse amines optimized for CO\textsubscript{2} capture.

\begin{table}[ht!]
    \centering
    \caption{Results of the SAGE-Amine Models for the Goal-Directed Benchmarks}
    \renewcommand{\arraystretch}{1.3} 
    \begin{tabular}{@{}c c c c c c c c c c@{}}
        \toprule
        \textbf{Task} & \textbf{Name} & \textbf{Amine Type} & \textbf{GA} & \textbf{LSTM} & \textbf{LSTM/GA} & \textbf{TD} & \textbf{TD/GA} & \textbf{XTD} & \textbf{XTD/GA} \\
        \specialrule{1.5pt}{0pt}{0pt}
        \multirow{10}{*}{Rediscovery} & MEA & Primary & 0.444  & 1.000  & 1.000  & 1.000  & 1.000  & 1.000  & 1.000 \\
        & AHMPD & Primary & 0.441  & 1.000  & 1.000  & 1.000  & 1.000  & 1.000  & 1.000 \\
        & DEA & Secondary & 0.600  & 1.000  & 1.000  & 1.000  & 1.000  & 1.000  & 1.000 \\
        & DA & Secondary & 0.522  & 1.000  & 1.000  & 1.000  & 1.000  & 1.000  & 1.000 \\
        & DEEA & Tertiary & 1.000  & 1.000  & 1.000  & 1.000  & 1.000  & 1.000  & 1.000 \\
        & MDEA & Tertiary & 1.000  & 1.000  & 1.000  & 1.000  & 1.000  & 1.000  & 1.000 \\
        & 2-MPZ & Cyclic & 1.000  & 1.000  & 1.000  & 1.000  & 1.000  & 1.000  & 1.000 \\
        & 2-PPE & Cyclic & 1.000  & 1.000  & 1.000  & 1.000  & 1.000  & 1.000  & 1.000 \\
        & HomoPZ & Cyclic & 0.567  & 1.000  & 1.000  & 1.000  & 1.000  & 1.000  & 1.000 \\
        & DETA & Poly & 0.600  & 1.000  & 1.000  & 1.000  & 1.000  & 1.000  & 1.000 \\
        \cmidrule(lr){1-10}
        \multirow{10}{*}{Similarity} & AMP & Primary & 0.777  & 0.817  & 0.817  & 0.817  & 0.817  & 0.815  & 0.817 \\
        & IPA & Primary & 0.617  & 0.790  & 0.790  & 0.790  & 0.790  & 1.000  & 0.790 \\
        & IPAP & Secondary & 0.796  & 0.842  & 0.842  & 0.842  & 0.842  & 0.837  & 0.842 \\
        & 4DMA1B & Tertiary & 0.814  & 0.851  & 0.851  & 0.851  & 0.851  & 0.731  & 0.851 \\
        & 1DMA2P & Tertiary & 0.768  & 0.828  & 0.828  & 0.828  & 0.828  & 0.831  & 0.828 \\
        & 1-MPZ & Cyclic & 0.672  & 0.812  & 0.812  & 0.807  & 0.807  & 0.806  & 0.808 \\
        & EPZ & Cyclic & 0.634  & 0.896  & 0.896  & 0.895  & 0.894  & 0.896  & 0.895 \\
        & PZ & Cyclic & 0.377  & 0.766  & 0.766  & 0.766  & 0.764  & 0.766  & 0.776 \\
        & AEP & Cyclic & 0.687  & 0.906  & 0.906  & 0.904  & 0.904  & 0.863  & 0.911 \\
        & TETA & Poly & 0.739  & 0.930  & 0.930  & 0.930  & 0.939  & 0.943  & 1.000 \\
        \cmidrule(lr){1-10}
        \multirow{3}{*}{\shortstack{Median \\Similarity}} & \multicolumn{2}{c}{DEAE-EO and 1M-2PPE} & 0.277  & 0.395  & 0.395  & 0.395  & 0.395  & 0.313  & 0.395 \\
        & \multicolumn{2}{c}{DEAE-EO and 1-(2HE)PP} & 0.335  & 0.398  & 0.398  & 0.398  & 0.398  & 0.364  & 0.395 \\
        & \multicolumn{2}{c}{1M-2PPE and 1-(2HE)PP} & 0.428  & 0.462  & 0.462  & 0.438  & 0.441  & 0.431  & 0.437 \\
        \cmidrule(lr){1-10}
        \multirow{4}{*}{Isomers} & \multicolumn{2}{c}{C\textsubscript{4}H\textsubscript{11}NO} & 0.997  & 1.000  & 1.000  & 1.000  & 1.000  & 1.000  & 1.000 \\
        & \multicolumn{2}{c}{C\textsubscript{4}H\textsubscript{11}NO\textsubscript{2}} & 0.903  & 1.000  & 1.000  & 1.000  & 1.000  & 1.000  & 1.000 \\
        & \multicolumn{2}{c}{C\textsubscript{5}H\textsubscript{12}N\textsubscript{2}} & 0.907  & 0.999  & 1.000  & 1.000  & 1.000  & 1.000  & 1.000 \\
        & \multicolumn{2}{c}{C\textsubscript{6}H\textsubscript{15}NO} & 0.920  & 1.000  & 1.000  & 1.000  & 1.000  & 1.000  & 1.000 \\
        \cmidrule(lr){1-10}
        \multicolumn{3}{c}{Total} & 18.821  & 23.692  & 23.693  & 23.660  & 23.671  & 23.596  & 23.744 \\
        \bottomrule
    \end{tabular}
    \label{tab:table_2}
\end{table}

\subsection{Goal-directed Benchmarks for SAGE-Amine Evaluation}
In the previous phase, the NLP models generated molecules without specific objectives, focusing on validating their status as real amines and analyzing the types produced. To push this further, goal-directed benchmarks were prepared to evaluate how well the models could generate amines that meet specific targets, as outlined in Table~\ref{tab:table_2}. This shift from unguided generation to targeted design allows for a more precise assessment of the model's ability to create compounds tailored to particular criteria.

These benchmarks focused on four key tasks: rediscovery, similarity, median similarity, and isomer generation. Specifically, they aimed to evaluate how well the models could rediscover known amines, generate structurally similar molecules, create isomers with defined molecular formulas, and design amines that fall between two target structures.

For an evaluation, three approaches were employed: (1) using a genetic algorithm (GA) for chemical diversification, (2) using pre-trained NLP models (LSTM, TD, and XTD), and (3) combining the NLP models with GA to form hybrid models (LSTM/GA, TD/GA, and XTD/GA). Each of the seven models generated an equal number of molecules, which were validated as real chemicals and amines before scoring based on their performance in the benchmarks.

The target molecules were 23 amines known for their efficiency in CO\textsubscript{2} capture. The tasks included rediscovering these specific amines, creating structurally similar ones, generating isomers with defined molecular formulas, and producing amines that bridged two target molecules in terms of structure. In the rediscovery task, all models except GA performed perfectly, with GA scoring 7.174. In the similarity task, the XTD/GA model achieved the highest score (8.518), followed by XTD and LSTM/GA. For the median similarity task, LSTM/GA and LSTM achieved the best scores, while GA lagged. In the isomer generation task, all models except LSTM and GA scored the maximum of 4.000, with GA scoring the lowest. Overall, the XTD/GA model demonstrated the best performance across all benchmarks, achieving a total score of 23.744, followed by LSTM/GA and LSTM. This indicates that combining NLP models with GA yields better results in tasks requiring structural diversification and optimization. The goal-directed benchmarks highlight the strengths of each approach, particularly in challenging tasks such as amine rediscovery and isomer generation.

\subsection{Single Property Optimization with SAGE-Amine}
Based on the goal-directed benchmark results, the combination of NLP and GA approaches outperformed using NLP alone, making LSTM/GA, TD/GA, and XTD/GA in SAGE-Amine particularly effective for de novo amine design. In generating amines optimized for CO\textsubscript{2} capture using SAGE-Amine, several key factors must be considered, including high cyclic capacity, rapid absorption rate, low enthalpy of absorption, and high equilibrium temperature sensitivity \cite{ref_71}. Firstly, an effective amine should demonstrate high cyclic capacity, which includes both high absorption capacity and efficient desorption of CO\textsubscript{2}. The dissociation constant (pKa) is generally correlated with CO\textsubscript{2} absorption and cyclic capacity, with tertiary amines showing a strong correlation, while primary and secondary amines tend to exhibit less correlation \cite{ref_4}. Secondly, the amine should react rapidly with CO\textsubscript{2}. Higher pKa values are typically associated with faster reaction rates \cite{ref_71}. Thirdly, the amine should have low regeneration energy requirements. A lower heat of absorption is preferred to minimize the energy required for desorption. Additionally, low viscosity is beneficial in reducing energy consumption during processing. Finally, the vapor pressure of the amine should be low to minimize volatility, reducing solvent loss through evaporation, which in turn enhances cost-effectiveness.

Single property optimization (SPO) tasks were focused on optimizing three essential properties for efficient CO\textsubscript{2} absorption: basicity (pKa), viscosity, and vapor pressure. For predicting pKa, the MolGpKa model was employed, which achieved an RMSE of 0.47, an MAE of 0.29, and an R2 of 0.97 \cite{ref_40}, demonstrating high accuracy in predicting the basicity of amines. In contrast, the datasets for viscosity and vapor pressure were collected and are summarized in Table~\ref{tab:table_1}. Our QSPR models for these properties were specifically developed by combining molecular fingerprints with gradient boosting models, as well as molecular graphs with D-MPNN. To identify the optimal hyperparameters, a grid search was conducted with 5-fold cross-validation, with the explored hyperparameters summarized in Table~\ref{tab:table_s2}. The performance metrics for the best models, selected based on the product of the R² scores from the training and cross-validation sets, are detailed in Tables~\ref{tab:table_s3}-~\ref{tab:table_s5}, with the metrics of the final models presented in Table~\ref{tab:table_s6}. As a result, the QSPR model for predicting viscosity, using rDesc/LGBM, achieved an R2 of 0.9605 and an MAE of 0.0468 on the test set. Similarly, the vapor pressure model, using Graph/D-MPNN, performed with an R2 of 0.9952 and an MAE of 0.2016 on the test set.

These three QSPR models were integrated into SAGE-Amine to create three SPO tasks aimed at either increasing pKa, reducing viscosity, or lowering the vapor pressure of amines. Additionally, three different types of restrictions on the amine generation process were applied: generating only primary and secondary amines, generating only tertiary, cyclic, and polyamines, and generating amines without any restrictions. Using three different models (LSTM/GA, TD/GA, and XTD/GA), amines were generated over 100 steps for each SPO task. The results of these SPO tasks are summarized in Table~\ref{tab:table_3}.

The SPO tasks for generating amines with high pKa values were carried out using three different models, with three amine-type restrictions. The results are illustrated in Figure S1. First, the models were used to primarily generate primary and secondary amines, as shown in Figure S1A. The LSTM/GA model achieved a median pKa value of 10 by the 3rd step, and values of 12 and 14 by the 7th and 12th steps, respectively. The TD/GA model generated amines with median pKa values above 10, 12, and 14 at the 3rd, 9th, and 30th steps. The XTD model surpassed pKa values of 10 and 12 at the 10th and 27th steps but did not exceed a median pKa of 14. Based on maximum values, the LSTM/GA, TD/GA, and XTD/GA models predicted amines with pKa values of 17.338, 15.177, and 14.034, respectively.

\begin{table}[ht!]
    \centering
    \caption{Results of the SAGE-Amine Models for the Property Optimization Tasks}
    \renewcommand{\arraystretch}{1.3} 
    \begin{tabular}{@{}c c c c c c@{}}
        \toprule
        \textbf{Task} & \textbf{Objective} & \textbf{Amine Type} & \textbf{LSTM/GA} & \textbf{TD/GA} & \textbf{XTD/GA} \\
        \specialrule{1.5pt}{0pt}{0pt}
        \multirow{9}{*}{\shortstack{Single\\Property\\Optimization}} & \multirow{3}{*}{\shortstack{High\\pKa}} & Primary and Secondary & 17.338  & 15.177  & 14.034 \\
        & & Tertiary, Cyclic, and Poly & 20.344  & 19.453  & 16.137 \\
        & & No restriction & 16.475  & 14.241  & 15.599 \\
        \cmidrule(lr){2-6}
        & \multirow{3}{*}{\shortstack{Low\\Viscosity}} & Primary and Secondary & -0.753  & -0.764  & -0.668 \\
        & & Tertiary, Cyclic, and Poly & -0.838  & -0.832  & -0.830 \\
        & & No restriction & -0.838  & -0.841  & -0.844 \\
        \cmidrule(lr){2-6}
        & \multirow{3}{*}{\shortstack{Low\\Vapor\\Pressure}} & Primary and Secondary & -25.774  & -25.898  & -24.001 \\
        & & Tertiary, Cyclic, and Poly & -25.750  & -26.151  & -25.877 \\
        & & No restriction & -25.922  & -26.151  & -25.878 \\
        \cmidrule(lr){1-6}
        \multirow{3}{*}{\shortstack{Multiple\\Property\\Optimization}} & \multirow{3}{*}{\shortstack{High\\CO\textsubscript{2}\\Absorption}} & Primary and Secondary & 0.870 & 0.873 & 0.921 \\
        & & Tertiary, Cyclic, and Poly & 0.858 & 0.909 & 0.878 \\
        & & No restriction & 0.863 & 0.868 & 0.914 \\
        \bottomrule
    \end{tabular}
    \label{tab:table_3}
\end{table}

Similarly, the models were tasked with generating tertiary, cyclic, and polyamines, which are shown in Figure S1B. The LSTM/GA model produced amines with median pKa values above 10, 12, and 14 at the 3rd, 7th, and 8th steps. The TD/GA achieved this at the 3rd, 7th, and 34th steps, while the XTD model did so at the 3rd, 21st, and 21st steps. For maximum values, the LSTM/GA, TD/GA, and XTD/GA models generated amines with predicted pKa values of 20.344, 19.453, and 16.137, respectively. Finally, when no restrictions were applied to the type of amine generated, the LSTM/GA model produced amines with median pKa values above 10, 12, and 14 at the 2nd, 8th, and 12th steps. The TD/GA exceeded 10 and 12 at the 3rd and 7th steps but did not surpass 14. The XTD model exceeded pKa values of 10, 12, and 14 at the 3rd, 18th, and 25th steps. In terms of maximum values, the LSTM/GA, TD/GA, and XTD/GA models generated amines with predicted pKa values of 16.475, 14.241, and 15.599, respectively. In summary, all three models were able to generate amines with predicted pKa values above 14, confirming that SAGE-Amine is effective at producing amines with high pKa values.

Similarly, the SPO task was performed to reduce viscosity, and the results are shown in Figure S2. First, the models were used to primarily generate primary and secondary amines with lower viscosity, as illustrated in Figure S2A. The LSTM/GA model produced amines with a viscosity below 0.5 by the 4th step, and values of 0 and -0.5 at the 6th and 47th steps. The TD/GA model reached these values at the 4th, 6th, and 63rd steps, while the XTD/GA model achieved 0.5 and 0 viscosity at the 6th and 10th steps but did not reach -0.5. Looking at the minimum values, the models generated amines predicted to have viscosities of -0.753, -0.764, and -0.668, respectively.

Next, the task was to generate tertiary, cyclic, and polyamines, as shown in Figure S2B. The LSTM/GA model achieved viscosities of 0.5, 0, and -0.5 at the 3rd, 5th, and 17th steps, while the TD/GA achieved this at the 3rd, 6th, and 19th steps, and the XTD/GA at the 3rd, 7th, and 50th steps. For minimum values, the LSTM/GA, TD/GA, and XTD/GA models produced amines with viscosities of -0.838, -0.832, and -0.830, respectively. Finally, when no restrictions were applied to the type of amine generated, the results are shown in Figure S2C. The LSTM/GA model generated amines with viscosities below 0.5, 0, and -0.5 by the 3rd, 5th, and 19th steps. The TD/GA achieved this at the 3rd, 6th, and 19th steps, and the XTD at the 2nd, 5th, and 17th steps. In terms of minimum values during the 100-step SPO task, the LSTM/GA, TD/GA, and XTD/GA models generated amines with predicted viscosities of -0.838, -0.841, and -0.844, respectively. In conclusion, all three models effectively generated amines with viscosities lower than -0.5, demonstrating that SAGE-Amine can successfully design amines with reduced viscosity.

Lastly, SPO tasks were performed to reduce the vapor pressure of amines, and the results are shown in Figure S3. First, the models were used to generate primary and secondary amines with lower vapor pressure, as seen in Figure S3A. The LSTM/GA model produced amines with predicted vapor pressure values of -5, -10, and -20 at the 1st, 3rd, and 6th steps, based on the median values. The TD/GA model reached these values at the 4th, 6th, and 13th steps, while the XTD/GA model achieved this at the 7th, 9th, and 28th steps. Looking at the minimum values over the 100-step SPO task, the LSTM/GA, TD/GA, and XTD/GA models generated amines with vapor pressures of -25.744, -25.898, and -24.001, respectively.

Next, the task was to generate tertiary, cyclic, and polyamines, as shown in Figure S3B. The LSTM/GA model generated amines with predicted vapor pressures of -5, -10, and -20 at the 1st, 3rd, and 6th steps, based on median values. The TD/GA achieved these values at the 1st, 2nd, and 7th steps, while the X-TD/GA did so at the 1st, 2nd, and 9th steps. Looking at the minimum values, the LSTM/GA, TD/GA, and XTD/GA models created amines with vapor pressures of -25.750, -26.151, and -25.877, respectively. Finally, when generating amines without restrictions on type, the results are shown in Figure S3C. The LSTM/GA model produced amines with predicted vapor pressures of -5, -10, and -20 at the 1st, 3rd, and 5th steps, based on median values. The TD/GA achieved this at the 1st, 2nd, and 7th steps, and the XTD/GA at the 1st, 2nd, and 8th steps. For the minimum values, the LSTM/GA, TD/GA, and XTD/GA models produced amines with vapor pressures of -25.922, -26.151, and -25.878, respectively. In conclusion, all three models successfully generated amines with vapor pressures below -20, demonstrating that SAGE-Amine effectively reduces vapor pressure in amine design.

In summary, the SPO tasks for generating amines with high pKa values, lower viscosity, and reduced vapor pressure demonstrate the effectiveness of SAGE-Amine in producing optimized molecules. Notably, SAGE-Amine’s generative capabilities extend beyond the limits of the experimental database used to develop the QSPR models, pushing into unexplored regions. This extrapolative exploration allowed SAGE-Amine to generate amines predicted to exceed the maximum pKa values and achieve minimum values for viscosity and vapor pressure beyond the boundaries of the original dataset. These results emphasize the strength of SAGE-Amine in discovering novel amines with properties that surpass the constraints of the QSPR training data, offering a powerful tool for amine design in various applications.

\begin{figure}[ht!]
  \centering
  \includegraphics[width=1.0\textwidth]{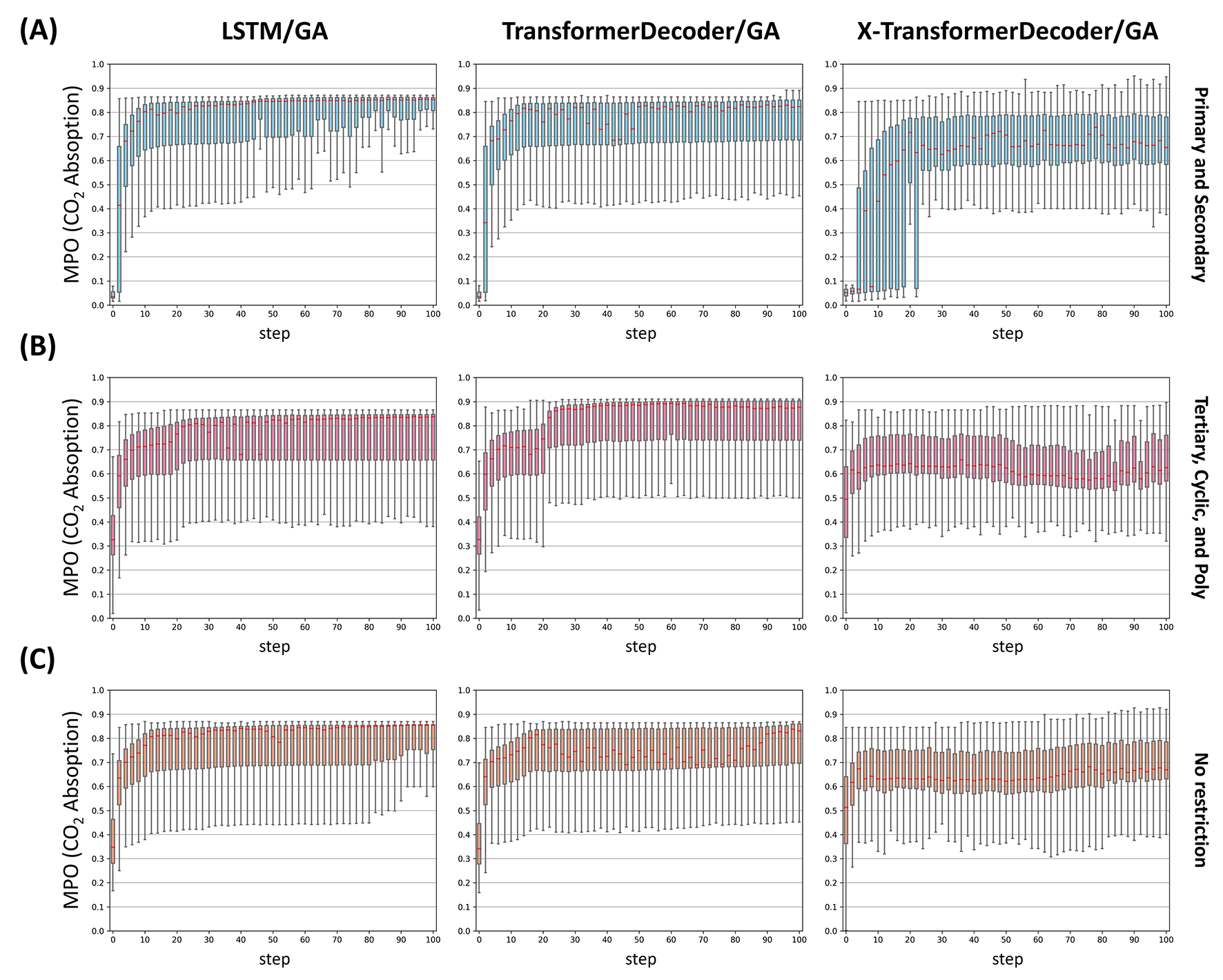}
  \caption{Multiple-Property Optimization Tasks for CO\textsubscript{2} Absorption of Amines. (A) Boxplots illustrate the SAGE-Amine process for optimizing multiple CO\textsubscript{2} absorption properties by generating primary and secondary amines using LSTM/GA, TD/GA, and XTD/GA methods. (B) Boxplots show the generation of tertiary, cyclic, and polyamines for multi-property optimization in CO\textsubscript{2} absorption. (C) Boxplots represent the steps for generating amines without restrictions on amine type. The median in each boxplot is highlighted in red.}
  \label{fig:fig_2}
\end{figure}

\subsection{Multiple Property Optimization with SAGE-Amine}
Based on the results of the SPO tasks, SAGE-Amine demonstrated its ability to optimize amines by modulating specific single properties and successfully identifying novel candidates with desired characteristics. However, the discovery of novel amines for CO\textsubscript{2} capture demands the simultaneous optimization of multiple properties rather than focusing on a single property. This necessitates the application of multiple-property optimization (MPO). In addition to the properties addressed in the SPO tasks, the CO\textsubscript{2} absorption process must also be economically viable. This requires that the amines be readily synthesizable, cost-effective, water-soluble, and exhibit favorable boiling and melting points that do not impede the cyclic absorption process. To address these requirements, the MPO tasks were prepared to evaluate its potential for discovering amines tailored for CO\textsubscript{2} capture. The evaluation included multiple objectives, including high pKa, low viscosity, low vapor pressure, high synthesizability, affordability, high aqueous solubility, and moderate boiling and melting points. For this purpose, pre-defined criteria were applied, followed by min-max scaling to normalize the objectives. The MPO score was subsequently defined as the average of these scaled objectives, enabling a comprehensive assessment of SAGE-Amine's performance in meeting the diverse requirements for CO\textsubscript{2} capture applications.

\begin{figure}[ht!]
  \centering
  \includegraphics[width=1.0\textwidth]{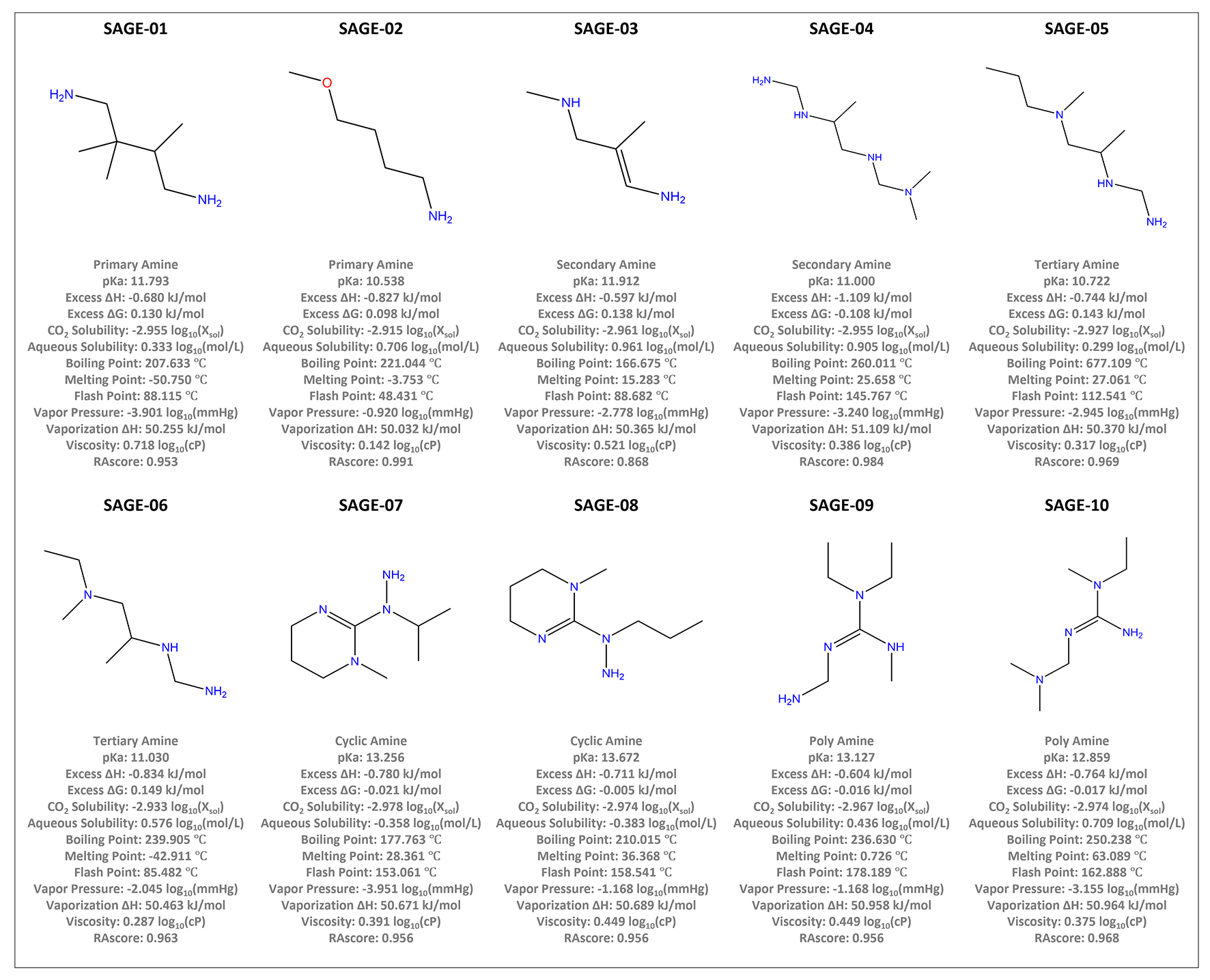}
  \caption{Top-ranked Amines Generated by SAGE-Amine. Top-ranked amines generated by SAGE-Amine and their predicted properties are shown. Carbon, nitrogen, and oxygen atoms are colored black, blue, and red, respectively.}
  \label{fig:fig_3}
\end{figure}

As with the SPO tasks, the MPO tasks for generating amines with multiple desired properties were performed using the LSTM/GA, TD/GA, and XTD/GA models under four amine restriction categories. The results are depicted in Figure~\ref{fig:fig_2} and summarized in Table~\ref{tab:table_3}. First, the generation of primary and secondary amines with higher MPO scores was carried out using iterative fine-tuning with SAGE-Amine, as illustrated in Figure~\ref{fig:fig_2}A. The LSTM/GA model achieved an MPO score exceeding 0.5 by the 3rd step and reached scores of 0.6, 0.7, and 0.8 at the 4th, 6th, and 11th steps, respectively. The TD/GA model attained these scores at the 3rd, 4th, 8th, and 13th steps. In contrast, the XTD/GA model reached scores of 0.5, 0.6, and 0.7 at the 12th, 17th, and 19th steps, respectively, but did not surpass 0.8 at the median level. For maximum MPO scores over the 100-step MPO task, the LSTM/GA, TD/GA, and XTD/GA models generated amines with scores of 0.870, 0.873, and 0.921, respectively.

Next, the task involved generating tertiary, cyclic, and polyamines, as shown in Figure~\ref{fig:fig_2}B. The LSTM/GA model achieved MPO scores of 0.5, 0.6, 0.7, and 0.8 at the 2nd, 3rd, 8th, and 23rd steps, respectively. The TD/GA model reached these scores at the 2nd, 3rd, 7th, and 22nd steps, while the XTD/GA achieved a score of 0.5 at the 2nd step but did not exceed 0.7 or 0.8 at the median level. In terms of maximum MPO scores, the LSTM/GA, TD/GA, and XTD/GA models achieved values of 0.858, 0.909, and 0.878, respectively. Finally, when no restrictions were applied to the type of amine generated, the results are presented in Figure~\ref{fig:fig_2}C. The LSTM/GA model produced amines with MPO scores exceeding 0.5, 0.6, 0.7, and 0.8 by the 2nd, 3rd, 5th, and 12th steps, respectively. Similarly, the TD/GA achieved these scores at the 2nd, 3rd, 5th, and 17th steps, whereas the XTD/GA reached scores of 0.5 and 0.6 at the 1st and 2nd steps but failed to exceed 0.7 or 0.8 at the median level. Regarding maximum MPO scores during the 100-step task, the LSTM/GA, TD/GA, and XTD/GA models produced amines with predicted viscosities of 0.863, 0.868, and 0.914, respectively.

To design amines with higher MPO scores for CO\textsubscript{2} capture, the SAGE-Amine models generated 384k primary, 1.21M secondary, 168k tertiary, 869k cyclic, and 2.13M polyamines in nine MPO tasks. The top 300 molecules per class were selected based on MPO scores and their six thermodynamic properties (CO\textsubscript{2} Henry’s law constant, CO\textsubscript{2} solubility, vaporization $\Delta$H, excess $\Delta$H, excess $\Delta$G, and flash point) were predicted via COSMO-RS simulations in a water-amine mixture (80 wt$\%$ water and 20 wt$\%$ amine). For comparison, the top two amines from each class were analyzed alongside 23 reference amines from the goal-directed benchmark, with predicted properties summarized in Tables~\ref{tab:table_s7}–~\ref{tab:table_s8} and the top 10 illustrated in Figure~\ref{fig:fig_3}. Compared to the reference median, top-ranked amines showed an increased pKa (+2.853) and boiling point (+86.248), decreased viscosity (-0.237), vapor pressure (-1.529), and melting point (-7.043), while synthetic accessibility, price, and aqueous solubility remaining similar. This indicates that SAGE-Amine effectively guided molecule generation towards the intended objectives. COSMO-RS results further showed a reduced CO\textsubscript{2} Henry’s law constant (-9.222), increased flash point (+9.903), slightly lower vaporization $\Delta$H (-0.280), and comparable CO\textsubscript{2} solubility, excess $\Delta$H, and excess $\Delta$G. Overall, the SAGE-Amine generated top-ranked amines competitive with references while achieving target objectives, highlighting the potential of generative models for task-specific amine discovery in CO\textsubscript{2} capture.

\section{Discussion}
The de novo design of molecules using generative models is emerging as a viable alternative to traditional methods such as virtual screening and combinatorial enumeration. In this study, the SAGE framework was extended to develop SAGE-Amine, tailored for amine generation in CO\textsubscript{2} absorption applications. By leveraging SMILES-based NLP models pre-trained on diverse amine structures, the SAGE-Amine enabled the autoregressive generation of novel amines with high validity, uniqueness, and diversity. Our findings suggest that different NLP architectures exhibit varying generative tendencies: LSTM and Transformer models primarily reproduced training set molecules, while TD and XTD generated a higher proportion of novel amines. Incorporating GA further enhanced molecular diversity, with XTD/GA achieving the best performance in goal-directed benchmarks. Furthermore, amine properties were optimized using SAGE-Amine for SPO and MPO tasks. Through iterative fine-tuning, SAGE-Amine demonstrated a progressive improvement in selecting high-scoring amines with desired characteristics such as high basicity, low viscosity, and optimal CO\textsubscript{2} absorption properties. These results underscore SAGE-Amine's potential for task-specific molecular discovery and its applicability in amine-based CO\textsubscript{2} capture.

This study demonstrated that SAGE-Amine can generate amines with desired properties, but there remains significant room for improvement in this generative approach. First, although the rediscovery task in the goal-directed benchmark suggests that the chemical space SAGE-Amine can explore is vast, the search space narrows due to the reliance on QSPR models for selecting top-ranked amines. To overcome this, it is essential to improve the prediction performance and application domain of QSPR models, as accurate predictions and broad applicability are crucial. Second, the CO\textsubscript{2} absorption MPO task targeted in this study still requires consideration of additional properties beyond those used in this work. These include the CO\textsubscript{2} cyclic capacity, CO\textsubscript{2} absorption rate, heat of reaction with CO\textsubscript{2}, corrosion rate, biodegradability, stability, and toxicity. Although the MPO tasks were defined for CO\textsubscript{2} absorption with the available QSPR models, employing a wider range of QSPR models to cover a broader application domain and more desired properties could lead to the discovery of better amines for CO\textsubscript{2} absorption. Third, the NLP models used in this study for pre-training and autoregressive generation were built using a single GPU. The results from large language models (LLMs) indicate that more model parameters and larger pre-training datasets enhance performance \cite{ref_72}. The amine datasets used for pre-training in this study were limited compared to the broader scope of GDB-13, and the NLP models had significantly fewer parameters than other LLMs. Future research should utilize NLP models with more parameters and larger pre-training datasets to achieve better performance.

Efficient CO\textsubscript{2} capture remains a major challenge in combating climate change, requiring improvements in amine-based solvents beyond conventional trial-and-error methods. This study presents SAGE-Amine, a generative framework that combines autoregressive NLP models with QSPR-based optimization to design novel amines for CO\textsubscript{2} absorption. By exploring beyond predefined chemical spaces, SAGE-Amine generates amines with enhanced properties, including higher pKa, lower viscosity, and reduced vapor pressure. The results highlight generative modeling as a powerful tool for accelerating material discovery and optimizing amines for multi-objective criteria. By shifting from traditional screening to de novo molecular design, SAGE-Amine enables the rapid discovery of task-specific chemicals, reducing reliance on costly, time-intensive synthesis. As generative models advance, they are poised to drive the next generation of chemical and material innovation.

\section{Author Contributions}
H.L. conceptualized and wrote the manuscript. H.C. and J.K. supported making quantitative structure-property relationship models. All authors reviewed the manuscript.

\section{Acknowledgments}

This research was supported by Quantum Advantage challenge research based on Quantum Computing through the National Research Foundation of Korea (NRF) funded by the Ministry of Science and ICT (RS-2023-00257288). H.L. acknowledges the Computational Systems Biology Laboratory at Yonsei University and Kyoung Tai No for providing access to the COSMOtherm software.

\section{Declarations}

\textbf{Competing interests \\ }The authors declare no competing interest.

\textbf{Funding}\\
H.L., H.C., and J.K. are financially supported by Quantum Advantage challenge research based on Quantum Computing through the National Research Foundation of Korea (NRF) funded by the Ministry of Science and ICT (RS-2023-00257288)..

\textbf{Code Availability \\ }All results in this work can be found at \href{https://github.com/hclim0213/HQNN-Amine}{https://github.com/hclim0213/SAGE-Amine}.



\appendix
\clearpage
\thispagestyle{empty}
\renewcommand{\thetable}{S\arabic{table}}
\renewcommand{\thefigure}{S\arabic{figure}}
\setcounter{table}{0}
\setcounter{figure}{0}

\section*{Supporting Information for}

\title{SAGE-Amine: Generative Amine Design with Multiple-Property Optimization for Efficient CO\textsubscript{2} Capture}
\maketitle

\author[a]{Hocheol Lim$^{\ast}$}
\author[b]{Hyein Cho$^{}$}
\author[a]{Jeonghoon Kim$^{}$}

\begin{flushleft}
    $^{\ast}$ Corresponding author: Hocheol Lim (ihc0213@yonsei.ac.kr)
\end{flushleft}

\vspace{0.5cm}

\textbf{\hspace{36pt}}The supporting information for ‘SAGE-Amine: Generative Amine Design with Multiple-Property Optimization for Efficient CO\textsubscript{2} Capture’ includes Figures S1-S3 for single property optimization results for high pKa, low viscosity, and low vapor pressure, respectively. It also includes Table S1 for performance metrics of the pre-trained models, Table S2 for the hyperparameter tuning procedure, Tables S3-S6 for performance metrics of QSPR models, and Table S7-S8 for predicted properties of amines used in this work.

\textbf{\hspace{36pt}}In this study, there are many abbreviations as follows. 1-(2HE)PP; 1-(2-Hydroxyethyl)piperdine, 1DMA2P; 1-Dimethylamino-2-Propanol, 1M-2PPE; 1-Methyl-2-piperidinemethanol, 1-MPZ; 1-Methylpiperazine, 2-MPZ; 2-Methylpiperazine, 2-PPE; 2-Piperidineethanol, 4DMA1B; 4-Dimethylamino-1-Butanol, AEP; 1-(2-Aminoethyl)Piperazine, AHMPD; 2-Amino-2-Hydroxymethyl-1,3-Propanediol, AMP; 2-Amino-2-Methyl-1-Propanol, BP; Becke–Perdew, CCUS; Carbon Capture, Utilization, and Storage, COSMO-RS; Conductor-like Screening Model for Real Solvents, DA; Diethylamine, DEA; Diethanolamine, DEAE-EO; 2-(2-Diethylaminoethoxy)ethanol, DEEA; N,N-Diethylethanolamine, DETA; Diethylenetriamine, DFT; density functional theory, D-MPNN; Directed message-passing neural network, ECFP; Extended-connectivity fingerprint, EPZ; 1-Ethyl-Piperazine, FCFP; Functional-class fingerprint, GA; Genetic Algorithm, HomoPZ; Homopiperazine, IPA; Isopropylamine, IPAP; 3-(Isopropylamino)Propanol, LGBM; Light Gradient Boosting Machine, LLM; Large Language Model, LSTM; Long-Short-Term Memory, MACCS; Molecular Access System, MDEA; Methyldiethanolamine, MEA; Monoethanolamine, MPO; Multiple Property Optimization, NLP; Natural Language Processing, PCFP; PubChem Fingerprint, PZ; Piperazine, QSAR; Quantitative Structure-Property Relationship, QSPR; Quantitative Structure-Activity Relationship, RAscore; Retrosynthetic Accessibility score , RF; Random Forest, RI; Resolution of Identity, SAGE; Scoring-Assisted Generative Exploration, SEDiv; Sphere Exclusion Diversity, SMILES; Simplified Molecular Input Line Entry System, SPO; Single Property Optimization, TD; Transformer Decoder, TETA; Triethylenetetramine, VLE; Vapor-Liquid Equilibrium, XGB; Extreme Gradient Boosting, XTD; X-Transformer Decoder.

\newpage
\begin{figure}[ht!]
  \centering
  \includegraphics[width=1.0\textwidth]{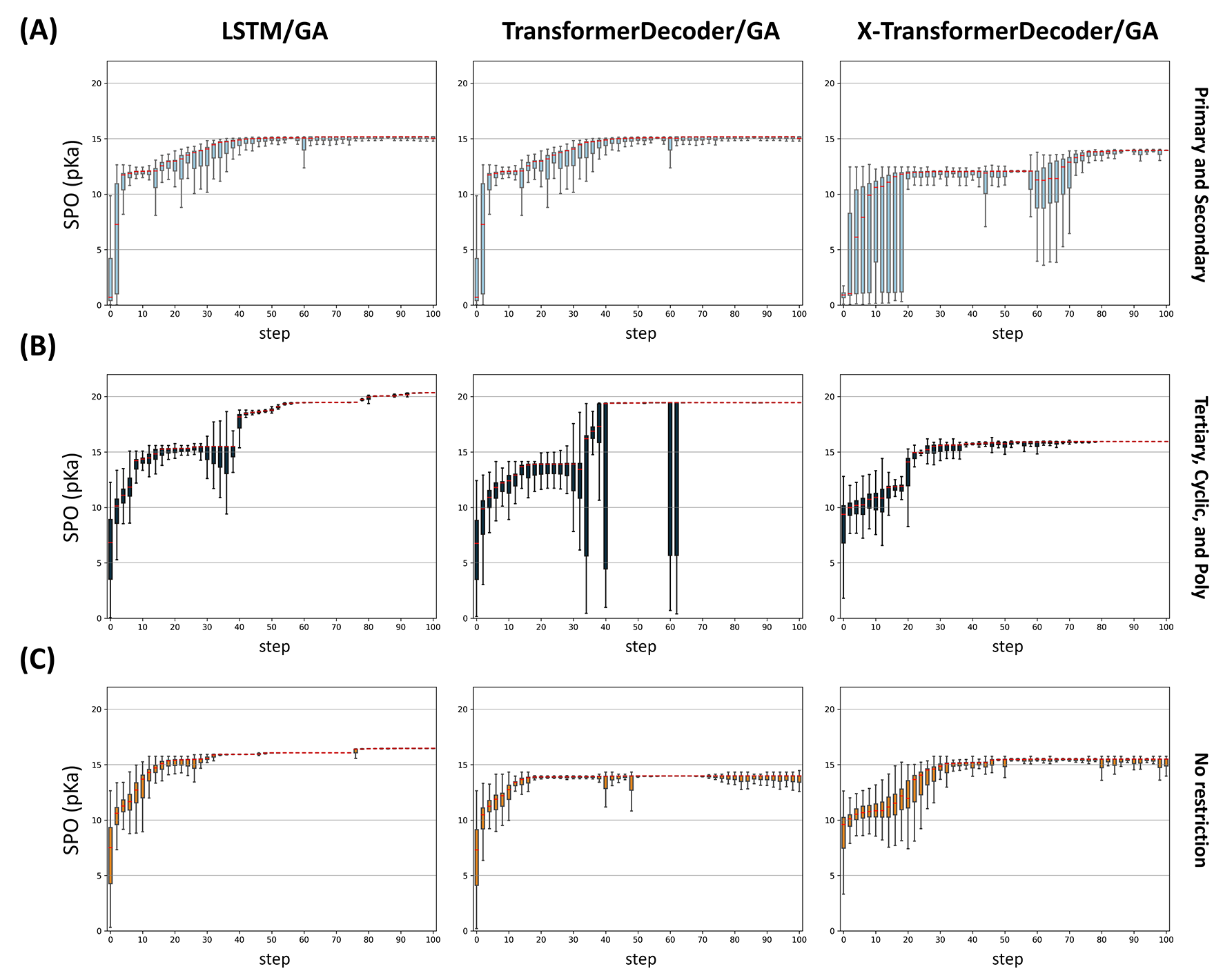}
  \label{fig:fig_s1}
  \caption{Single-Property Optimization Tasks for Low Viscosity of Amines. (A) Boxplots illustrate the SAGE-Amine process for minimizing viscosity by generating primary and secondary amines using LSTM/GA, TD/GA, and XTD/GA methods. (B) Boxplots depict the generation of tertiary, cyclic, and polyamines for minimizing viscosity. (C) Boxplots represent the steps for generating amines without restrictions on amine type. The medians in each boxplot are highlighted in red.}
\end{figure}

\newpage
\begin{figure}[ht!]
  \centering
  \includegraphics[width=1.0\textwidth]{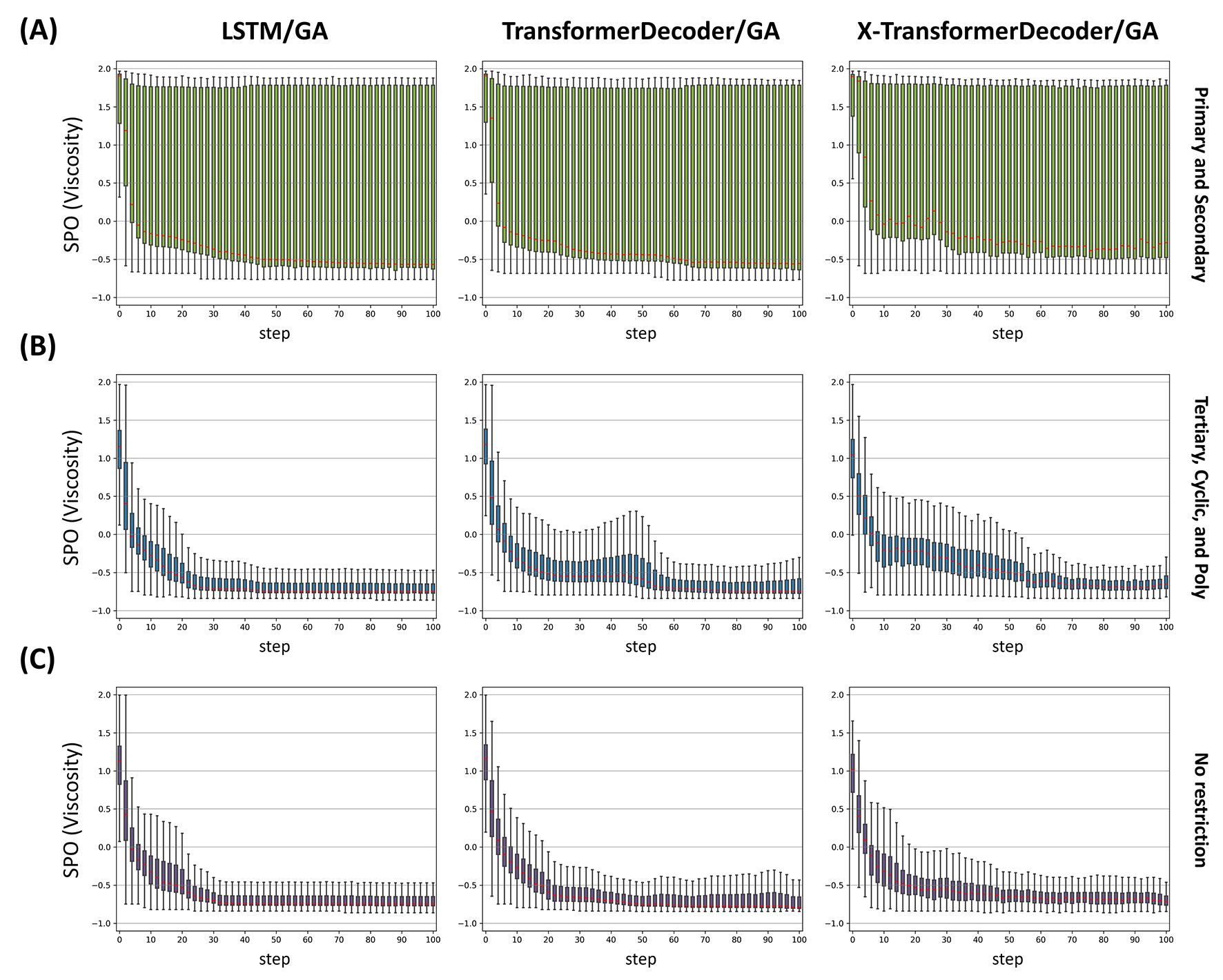}
  \label{fig:fig_s2}
  \caption{Single-Property Optimization Tasks for High pKa of Amines. (A) Boxplots show the SAGE-Amine process for maximizing pKa through the generation of primary and secondary amines using LSTM/GA, TD/GA, and XTD/GA methods. (B) Boxplots depict the process of generating tertiary, cyclic, and polyamines. (C) Boxplots illustrate the steps for generating amines without restrictions on amine type. The medians in each boxplot are marked in red.}
\end{figure}

\newpage
\begin{figure}[ht!]
  \centering
  \includegraphics[width=1.0\textwidth]{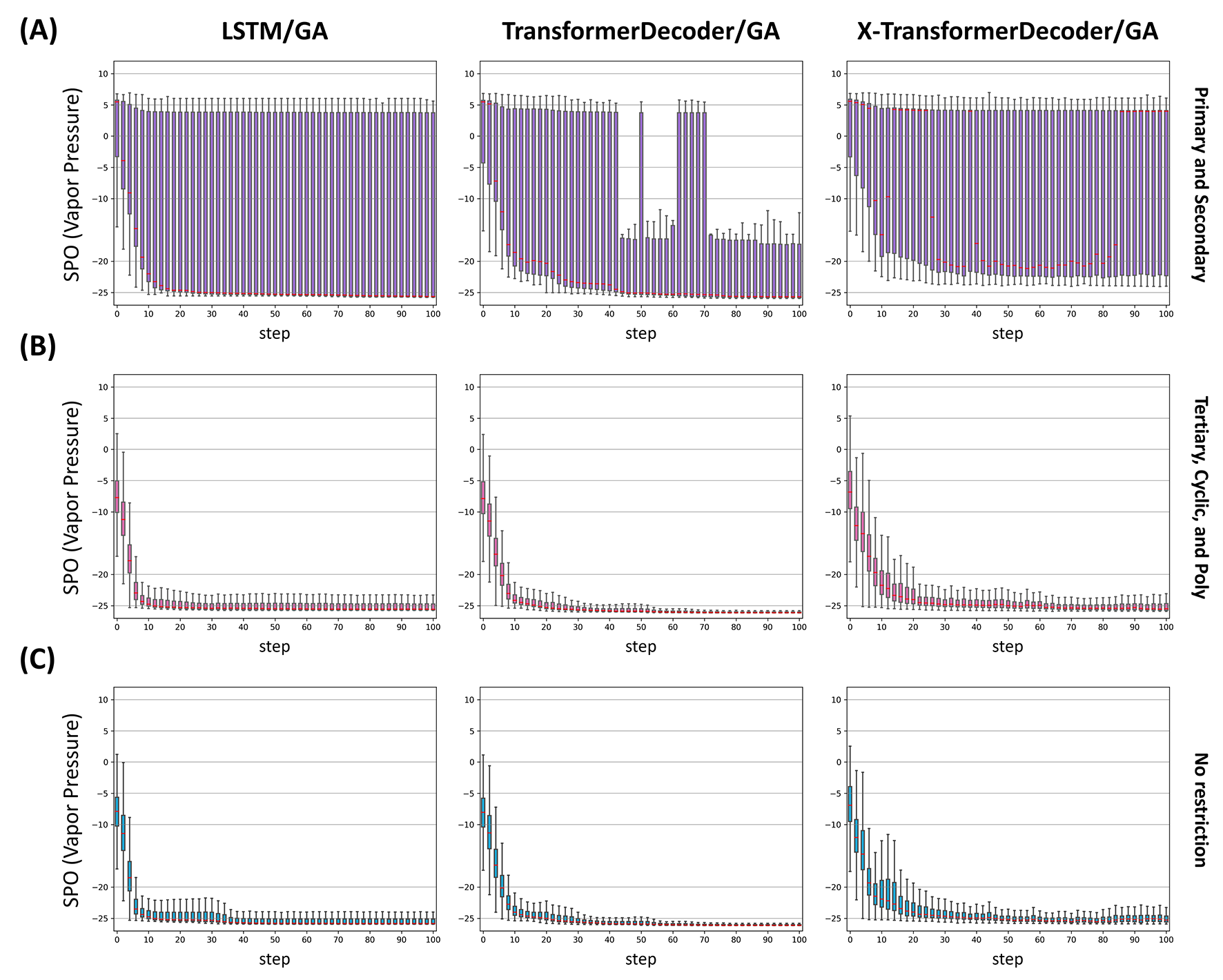}
  \label{fig:fig_s3}
  \caption{Single-Property Optimization Tasks for Low Vapor Pressure of Amines. (A) Boxplots illustrate the SAGE-Amine process for reducing vapor pressure by generating primary and secondary amines using LSTM/GA, TD/GA, and XTD/GA methods. (B) Boxplots show the generation of tertiary, cyclic, and polyamines aimed at minimizing vapor pressure. (C) Boxplots represent the generation of amines without restrictions on amine type. The medians in each boxplot are highlighted in red.}
\end{figure}

\newpage
\begin{sidewaystable}
    \centering
    \caption{Performance Metrics of Pre-trained Models in this work}
    \label{tab:table_s1}
    \vspace{5pt}
    \renewcommand{\arraystretch}{1.3}
    \begin{tabular}{c c c c c c c c c c c c c c}
        \toprule
        \textbf{Dataset} & \textbf{Models} & \textbf{Samples} & \textbf{Validity} & \textbf{Uniqueness} & \textbf{Novelty} & \textbf{SEDiv} & \textbf{Amine} & \textbf{Primary} & \textbf{Secondary} & \textbf{Tertiary} & \textbf{Cyclic} & \textbf{Poly} \\
        \specialrule{1.5pt}{0pt}{0pt}
        \multirow{8}{*}{Amine250} & \multirow{2}{*}{LSTM} & 5,000 & 0.963  & 0.977  & 0.502  & 0.537  & 0.977  & 0.068  & 0.104  & 0.037  & 0.666  & 0.103 \\
        & & 10,000 & 0.962  & 0.974  & 0.494  & 0.422  & 0.976  & 0.068  & 0.101  & 0.037  & 0.668  & 0.103 \\
        \cmidrule(lr){2-13}
        & \multirow{2}{*}{Transformer} & 5,000 & 0.972  & 0.618  & 0.003  & 0.387  & 0.972  & 0.064  & 0.106  & 0.033  & 0.670  & 0.099 \\
        & & 10,000 & 0.974  & 0.614  & 0.003  & 0.311  & 0.968  & 0.066  & 0.109  & 0.031  & 0.666  & 0.095 \\
        \cmidrule(lr){2-13}
        & \multirow{2}{*}{TD} & 5,000 & 0.962  & 0.962  & 0.748  & 0.517  & 0.958  & 0.051  & 0.092  & 0.026  & 0.694  & 0.095 \\
        & & 10,000 & 0.962  & 0.963  & 0.752  & 0.411  & 0.960  & 0.053  & 0.096  & 0.026  & 0.694  & 0.092 \\
        \cmidrule(lr){2-13}
        & \multirow{2}{*}{XTD} & 5,000 & 0.902  & 0.782  & 0.733  & 0.326  & 0.798  & 0.033  & 0.141  & 0.067  & 0.275  & 0.283 \\
        & & 10,000 & 0.904  & 0.766  & 0.722  & 0.260  & 0.803  & 0.031  & 0.140  & 0.069  & 0.271  & 0.291 \\
        \cmidrule(lr){1-13}
        \multirow{8}{*}{\shortstack{Amine250\\reduced}} & \multirow{2}{*}{LSTM} & 5,000 & 0.962  & 0.969  & 0.516  & 0.540  & 0.968  & 0.066  & 0.110  & 0.030  & 0.663  & 0.101 \\
        & & 10,000 & 0.963  & 0.965  & 0.514  & 0.425  & 0.968  & 0.065  & 0.108  & 0.034  & 0.658  & 0.103 \\
        \cmidrule(lr){2-13}
        & \multirow{2}{*}{Transformer} & 5,000 & 0.976  & 0.609  & 0.004  & 0.384  & 0.969  & 0.071  & 0.104  & 0.038  & 0.658  & 0.100 \\
        & & 10,000 & 0.977  & 0.607  & 0.003  & 0.309  & 0.967  & 0.070  & 0.099  & 0.037  & 0.666  & 0.096 \\
        \cmidrule(lr){2-13}
        & \multirow{2}{*}{TD} & 5,000 & 0.953  & 0.964  & 0.746  & 0.516  & 0.961  & 0.058  & 0.103  & 0.028  & 0.688  & 0.084 \\
        & & 10,000 & 0.956  & 0.963  & 0.748  & 0.410  & 0.961  & 0.059  & 0.102  & 0.027  & 0.686  & 0.086 \\
        \cmidrule(lr){2-13}
        & \multirow{2}{*}{XTD} & 5,000 & 0.850  & 0.772  & 0.593  & 0.276  & 0.976  & 0.184  & 0.003  & 0.000  & 0.627  & 0.162 \\
        & & 10,000 & 0.845  & 0.710  & 0.567  & 0.212  & 0.974  & 0.176  & 0.003  & 0.000  & 0.629  & 0.167 \\
        \cmidrule(lr){1-13}
        \multirow{8}{*}{Amine300} & \multirow{2}{*}{LSTM} & 5,000 & 0.960  & 0.976  & 0.765  & 0.579  & 0.973  & 0.030  & 0.095  & 0.017  & 0.738  & 0.094 \\
        & & 10,000 & 0.962  & 0.976  & 0.767  & 0.465  & 0.974  & 0.026  & 0.091  & 0.020  & 0.742  & 0.095 \\
        \cmidrule(lr){2-13}
        & \multirow{2}{*}{Transformer} & 5,000 & 0.978  & 0.620  & 0.002  & 0.400  & 0.978  & 0.025  & 0.109  & 0.032  & 0.709  & 0.103 \\
        & & 10,000 & 0.979  & 0.619  & 0.003  & 0.331  & 0.980  & 0.024  & 0.108  & 0.029  & 0.720  & 0.099 \\
        \cmidrule(lr){2-13}
        & \multirow{2}{*}{TD} & 5,000 & 0.961  & 0.976  & 0.841  & 0.525  & 0.971  & 0.018  & 0.092  & 0.018  & 0.729  & 0.114 \\
        & & 10,000 & 0.962  & 0.973  & 0.834  & 0.416  & 0.969  & 0.017  & 0.101  & 0.018  & 0.722  & 0.110 \\
        \cmidrule(lr){2-13}
        & \multirow{2}{*}{XTD} & 5,000 & 0.912  & 0.929  & 0.910  & 0.289  & 0.977  & 0.002  & 0.036  & 0.006  & 0.473  & 0.460 \\
        & & 10,000 & 0.915  & 0.890  & 0.872  & 0.224  & 0.977  & 0.002  & 0.038  & 0.008  & 0.473  & 0.457 \\
        \cmidrule(lr){1-13}
        \multirow{8}{*}{\shortstack{Amine300\\reduced}} & \multirow{2}{*}{LSTM} & 5,000 & 0.955  & 0.980  & 0.717  & 0.565  & 0.978  & 0.027  & 0.107  & 0.029  & 0.720  & 0.095 \\
        & & 10,000 & 0.958  & 0.977  & 0.717  & 0.456  & 0.976  & 0.028  & 0.109  & 0.027  & 0.718  & 0.094 \\
        \cmidrule(lr){2-13}
        & \multirow{2}{*}{Transformer} & 5,000 & 0.973  & 0.613  & 0.003  & 0.400  & 0.972  & 0.028  & 0.094  & 0.031  & 0.723  & 0.096 \\
        & & 10,000 & 0.971  & 0.610  & 0.003  & 0.332  & 0.973  & 0.030  & 0.098  & 0.029  & 0.717  & 0.100 \\
        \cmidrule(lr){2-13}
        & \multirow{2}{*}{TD} & 5,000 & 0.956  & 0.969  & 0.829  & 0.510  & 0.963  & 0.020  & 0.098  & 0.023  & 0.721  & 0.101 \\
        & & 10,000 & 0.956  & 0.967  & 0.822  & 0.405  & 0.962  & 0.019  & 0.097  & 0.023  & 0.718  & 0.105 \\
        \cmidrule(lr){2-13}
        & \multirow{2}{*}{XTD} & 5,000 & 0.547  & 0.986  & 0.983  & 0.603  & 0.987  & 0.009  & 0.064  & 0.014  & 0.837  & 0.062 \\
        & & 10,000 & 0.547  & 0.979  & 0.977  & 0.534  & 0.988  & 0.008  & 0.059  & 0.013  & 0.846  & 0.063 \\
        \bottomrule
    \end{tabular}
\end{sidewaystable}

\newpage
\begin{table}
    \centering
    \caption{Hyperparameters Used in Hyperparameter Tuning Procedure}
    \label{tab:table_s2}
    \vspace{5pt}
    \renewcommand{\arraystretch}{1.3}
    \begin{tabular}{ccc}
        \toprule
        \textbf{Method} & \textbf{Tuning Parameters} & \textbf{Fixed Parameters} \\
        \specialrule{1.5pt}{0pt}{0pt}
        \multirow{3}{*}{LGBM} & boosting$\_$type = gbdt, dart & \\
        & n$\_$estimators = 100, 500, 1000, 2000, 3000 & \\
        & learning$\_$rate = 0.01, 0.05, 0.1 & \\
        \cmidrule(lr){1-3}
        \multirow{2}{*}{RF} & n$\_$estimators = 100, 500, 1000, 2000, 3000 &  \\
        & max$\_$depth = 10, 20, 30 & \\
        \cmidrule(lr){1-3}
        \multirow{4}{*}{XGB} & booster = gbtree, dart & \\
        & n$\_$estimators = 100, 500, 1000, 2000, 3000 & \\
        & max$\_$depth = 10, 20, 30 & \\
        & learning$\_$rate = 0.01, 0.05, 0.1 & \\
        \cmidrule(lr){1-3}
        \multirow{4}{*}{D-MPNN} & hidden$\_$unit = 1024, 2048 & learning$\_$rate = 0.001 \\
        & step = 1, 2, 4 & dropout = 0.1 \\
        & & num$\_$epochs = 300 \\
        & & batch$\_$size = 128 \\
        \bottomrule
    \end{tabular}
\end{table}

\newpage
\begin{sidewaystable}[ht!]
    \centering
    \caption{Mean R-squared from 5-fold Cross-Validation Sets for QSPR Models in Property Optimization Tasks (continued)}
    \vspace{5pt}
    \label{tab:table_s3}
    \renewcommand{\arraystretch}{1.3}  
    \begin{tabular}{@{}cccccccc@{}}
        \toprule
        \multirow{2}{*}{\textbf{Task}} & \multirow{2}{*}{\textbf{Descriptor}} 
          & \multicolumn{3}{c}{\textbf{Train R-squared}} 
          & \multicolumn{3}{c}{\textbf{Test R-squared}} \\
          \cmidrule(lr){3-8}
         & & \textbf{RF} & \textbf{XGB} & \textbf{LGBM} & \textbf{RF} & \textbf{XGB} & \textbf{LGBM} \\
        \specialrule{1.5pt}{0pt}{0pt}
        
        \multirow{17}{*}{\textbf{Viscosity}} & Avalon & 0.9575 $\pm$ 0.0013 & 0.9533 $\pm$ 0.0009 & 0.9409 $\pm$ 0.0028 & 0.8237 $\pm$ 0.0215 & 0.8262 $\pm$ 0.0183 & 0.8459 $\pm$ 0.0135 \\
        \cmidrule(lr){2-8}
        & ECFP6 & 0.9692 $\pm$ 0.0009 & 0.9866 $\pm$ 0.0002 & 0.9853 $\pm$ 0.0003 & 0.8396 $\pm$ 0.0199 & 0.8774 $\pm$ 0.0154 & 0.8891 $\pm$ 0.0123 \\
        \cmidrule(lr){2-8}
        & Extended & 0.9577 $\pm$ 0.0022 & 0.9680 $\pm$ 0.0017 & 0.9526 $\pm$ 0.0021 & 0.8258 $\pm$ 0.0222 & 0.8459 $\pm$ 0.0231 & 0.8650 $\pm$ 0.0182 \\
        \cmidrule(lr){2-8}
        & FCFP4 & 0.9352 $\pm$ 0.0028 & 0.9445 $\pm$ 0.0033 & 0.8944 $\pm$ 0.0041 & 0.7421 $\pm$ 0.0355 & 0.7699 $\pm$ 0.0319 & 0.7573 $\pm$ 0.0227 \\
        \cmidrule(lr){2-8}
        & MACCS & 0.9599 $\pm$ 0.0013 & 0.9651 $\pm$ 0.0016 & 0.9512 $\pm$ 0.0013 & 0.8530 $\pm$ 0.0211 & 0.8610 $\pm$ 0.0216 & 0.8707 $\pm$ 0.0136 \\
        \cmidrule(lr){2-8}
        & Morgan & 0.9681 $\pm$ 0.0009 & 0.9897 $\pm$ 0.0005 & 0.9774 $\pm$ 0.0007 & 0.8402 $\pm$ 0.0190 & 0.8808 $\pm$ 0.0214 & 0.8726 $\pm$ 0.0138 \\
        \cmidrule(lr){2-8}
        & PCFP & 0.9793 $\pm$ 0.0010 & 0.9905 $\pm$ 0.0009 & 0.9857 $\pm$ 0.0009 & 0.8860 $\pm$ 0.0134 & 0.9044 $\pm$ 0.0104 & 0.9199 $\pm$ 0.0110 \\
        \cmidrule(lr){2-8}
        & rDesc & 0.9904 $\pm$ 0.0004 & 0.9994 $\pm$ 0.0000 & 0.9985 $\pm$ 0.0002 & 0.9345 $\pm$ 0.0078 & 0.9371 $\pm$ 0.0113 & 0.9605 $\pm$ 0.0060 \\
        \cmidrule(lr){2-8}
        & rPair & 0.9760 $\pm$ 0.0009 & 0.9994 $\pm$ 0.0000 & 0.9950 $\pm$ 0.0005 & 0.8892 $\pm$ 0.0125 & 0.9192 $\pm$ 0.0125 & 0.9199 $\pm$ 0.0142 \\
        \cmidrule(lr){2-8}
        & rTorsion & 0.8413 $\pm$ 0.0044 & 0.9231 $\pm$ 0.0046 & 0.8054 $\pm$ 0.0021 & 0.6810 $\pm$ 0.0265 & 0.7730 $\pm$ 0.0272 & 0.6508 $\pm$ 0.0162 \\
        \cmidrule(lr){2-8}
        & Standard & 0.9471 $\pm$ 0.0022 & 0.9561 $\pm$ 0.0014 & 0.9367 $\pm$ 0.0013 & 0.8152 $\pm$ 0.0259 & 0.8327 $\pm$ 0.0183 & 0.8457 $\pm$ 0.0141 \\
        \cmidrule(lr){2-8}
        & MACCS+Avalon & 0.9687 $\pm$ 0.0010 & 0.9736 $\pm$ 0.0009 & 0.9725 $\pm$ 0.0010 & 0.8709 $\pm$ 0.0156 & 0.8760 $\pm$ 0.0153 & 0.8929 $\pm$ 0.0097 \\
        \cmidrule(lr){2-8}
        & MACCS+ECFP6 & 0.9801 $\pm$ 0.0005 & 0.9912 $\pm$ 0.0005 & 0.9875 $\pm$ 0.0004 & 0.8980 $\pm$ 0.0154 & 0.9057 $\pm$ 0.0137 & 0.9196 $\pm$ 0.0088 \\
        \cmidrule(lr){2-8}
        & MACCS+Extended & 0.9713 $\pm$ 0.0013 & 0.9844 $\pm$ 0.0013 & 0.9712 $\pm$ 0.0010 & 0.8657 $\pm$ 0.0162 & 0.8817 $\pm$ 0.0156 & 0.8989 $\pm$ 0.0105 \\
        \cmidrule(lr){2-8}
        & MACCS+FCFP4 & 0.9621 $\pm$ 0.0011 & 0.9691 $\pm$ 0.0014 & 0.9575 $\pm$ 0.0015 & 0.8566 $\pm$ 0.0213 & 0.8645 $\pm$ 0.0229 & 0.8780 $\pm$ 0.0153 \\
        \cmidrule(lr){2-8}
        & MACCS+PCFP & 0.9820 $\pm$ 0.0010 & 0.9928 $\pm$ 0.0005 & 0.9892 $\pm$ 0.0008 & 0.9043 $\pm$ 0.0161 & 0.9136 $\pm$ 0.0132 & 0.9284 $\pm$ 0.0100 \\
        \cmidrule(lr){2-8}
        & MACCS+Standard & 0.9723 $\pm$ 0.0011 & 0.9840 $\pm$ 0.0010 & 0.9791 $\pm$ 0.0012 & 0.8739 $\pm$ 0.0180 & 0.8879 $\pm$ 0.0158 & 0.8977 $\pm$ 0.0099 \\
        \bottomrule
    \end{tabular}
\end{sidewaystable}

\newpage
\begin{sidewaystable}[ht!]
    \centering
    Table S3: Mean R-squared from 5-fold Cross-Validation Sets for QSPR Models in Property Optimization Tasks (continued)
    \vspace{5pt}
    \label{tab:table_s3_2}
    \renewcommand{\arraystretch}{1.3}  
    \begin{tabular}{@{}cccccccc@{}}
        \toprule
        \multirow{2}{*}{\textbf{Task}} & \multirow{2}{*}{\textbf{Descriptor}} 
          & \multicolumn{3}{c}{\textbf{Train R-squared}} 
          & \multicolumn{3}{c}{\textbf{Test R-squared}} \\
          \cmidrule(lr){3-8}
         & & \textbf{RF} & \textbf{XGB} & \textbf{LGBM} & \textbf{RF} & \textbf{XGB} & \textbf{LGBM} \\
        \specialrule{1.5pt}{0pt}{0pt}
        \multirow{17}{*}{\textbf{\shortstack{Boiling\\Point}}} & Avalon & 0.9222 $\pm$ 0.0025 & 0.9308 $\pm$ 0.0037 & 0.8755 $\pm$ 0.0074 & 0.6635 $\pm$ 0.0849 & 0.6425 $\pm$ 0.0930 & 0.7114 $\pm$ 0.0543 \\
        \cmidrule(lr){2-8}
        & ECFP6 & 0.8988 $\pm$ 0.0041 & 0.8410 $\pm$ 0.0059 & 0.7729 $\pm$ 0.0155 & 0.5122 $\pm$ 0.1357 & 0.5029 $\pm$ 0.1698 & 0.6394 $\pm$ 0.0762 \\
        \cmidrule(lr){2-8}
        & Extended & 0.8928 $\pm$ 0.0039 & 0.9218 $\pm$ 0.0058 & 0.7981 $\pm$ 0.0147 & 0.6296 $\pm$ 0.0928 & 0.6334 $\pm$ 0.1405 & 0.7020 $\pm$ 0.0608 \\
        \cmidrule(lr){2-8}
        & FCFP4 & 0.7374 $\pm$ 0.0223 & 0.7579 $\pm$ 0.0235 & 0.7419 $\pm$ 0.0229 & 0.6234 $\pm$ 0.0655 & 0.6376 $\pm$ 0.0775 & 0.6502 $\pm$ 0.0718 \\
        \cmidrule(lr){2-8}
        & MACCS & 0.8777 $\pm$ 0.0044 & 0.8954 $\pm$ 0.0068 & 0.7954 $\pm$ 0.0202 & 0.7389 $\pm$ 0.0260 & 0.6835 $\pm$ 0.1351 & 0.7087 $\pm$ 0.0629 \\
        \cmidrule(lr){2-8}
        & Morgan & 0.8943 $\pm$ 0.0038 & 0.8504 $\pm$ 0.0064 & 0.8072 $\pm$ 0.0113 & 0.5317 $\pm$ 0.1565 & 0.5412 $\pm$ 0.2078 & 0.6477 $\pm$ 0.0725 \\
        \cmidrule(lr){2-8}
        & PCFP & 0.9176 $\pm$ 0.0066 & 0.9677 $\pm$ 0.0017 & 0.8255 $\pm$ 0.0234 & 0.6621 $\pm$ 0.1039 & 0.6091 $\pm$ 0.1743 & 0.7705 $\pm$ 0.0770 \\
        \cmidrule(lr){2-8}
        & rDesc & 0.9731 $\pm$ 0.0016 & 0.9945 $\pm$ 0.0017 & 0.9759 $\pm$ 0.0073 & 0.8373 $\pm$ 0.0475 & 0.7925 $\pm$ 0.0411 & 0.8717 $\pm$ 0.0420 \\
        \cmidrule(lr){2-8}
        & rPair & 0.7527 $\pm$ 0.0333 & 0.8065 $\pm$ 0.0296 & 0.7711 $\pm$ 0.0319 & 0.6916 $\pm$ 0.1022 & 0.7176 $\pm$ 0.1009 & 0.7239 $\pm$ 0.1067 \\
        \cmidrule(lr){2-8}
        & rTorsion & 0.6716 $\pm$ 0.0323 & 0.7500 $\pm$ 0.0344 & 0.7181 $\pm$ 0.0337 & 0.5990 $\pm$ 0.0949 & 0.6705 $\pm$ 0.1067 & 0.6700 $\pm$ 0.1087 \\
        \cmidrule(lr){2-8}
        & Standard & 0.8217 $\pm$ 0.0248 & 0.8483 $\pm$ 0.0281 & 0.8179 $\pm$ 0.0277 & 0.6376 $\pm$ 0.0746 & 0.6952 $\pm$ 0.0992 & 0.7067 $\pm$ 0.0868 \\
        \cmidrule(lr){2-8}
        & MACCS+Avalon & 0.9398 $\pm$ 0.0033 & 0.9737 $\pm$ 0.0020 & 0.8789 $\pm$ 0.0076 & 0.7510 $\pm$ 0.0575 & 0.6890 $\pm$ 0.1643 & 0.7521 $\pm$ 0.0476 \\
        \cmidrule(lr){2-8}
        & MACCS+ECFP6 & 0.9483 $\pm$ 0.0032 & 0.9832 $\pm$ 0.0017 & 0.9633 $\pm$ 0.0022 & 0.7545 $\pm$ 0.0654 & 0.7826 $\pm$ 0.0432 & 0.7561 $\pm$ 0.0547 \\
        \cmidrule(lr){2-8}
        & MACCS+Extended & 0.9409 $\pm$ 0.0026 & 0.9721 $\pm$ 0.0018 & 0.8835 $\pm$ 0.0108 & 0.7571 $\pm$ 0.0518 & 0.7298 $\pm$ 0.1094 & 0.7626 $\pm$ 0.0570 \\
        \cmidrule(lr){2-8}
        & MACCS+FCFP4 & 0.9003 $\pm$ 0.0055 & 0.9182 $\pm$ 0.0069 & 0.8157 $\pm$ 0.0199 & 0.7719 $\pm$ 0.0266 & 0.7174 $\pm$ 0.1344 & 0.7420 $\pm$ 0.0636 \\
        \cmidrule(lr){2-8}
        & MACCS+PCFP & 0.9378 $\pm$ 0.0038 & 0.9803 $\pm$ 0.0017 & 0.8571 $\pm$ 0.0233 & 0.7058 $\pm$ 0.0842 & 0.6907 $\pm$ 0.1516 & 0.7944 $\pm$ 0.0729 \\
        \cmidrule(lr){2-8}
        & MACCS+Standard & 0.9380 $\pm$ 0.0019 & 0.9691 $\pm$ 0.0021 & 0.9144 $\pm$ 0.0084 & 0.7447 $\pm$ 0.0768 & 0.7132 $\pm$ 0.1486 & 0.7457 $\pm$ 0.0786 \\
        \bottomrule
    \end{tabular}
\end{sidewaystable}

\newpage
\begin{sidewaystable}[ht!]
    \centering
    Table S3: Mean R-squared from 5-fold Cross-Validation Sets for QSPR Models in Property Optimization Tasks (continued)
    \vspace{5pt}
    \label{tab:table_s3_3}
    \renewcommand{\arraystretch}{1.3}  
    \begin{tabular}{@{}cccccccc@{}}
        \toprule
        \multirow{2}{*}{\textbf{Task}} & \multirow{2}{*}{\textbf{Descriptor}} 
          & \multicolumn{3}{c}{\textbf{Train R-squared}} 
          & \multicolumn{3}{c}{\textbf{Test R-squared}} \\
          \cmidrule(lr){3-8}
         & & \textbf{RF} & \textbf{XGB} & \textbf{LGBM} & \textbf{RF} & \textbf{XGB} & \textbf{LGBM} \\
        \specialrule{1.5pt}{0pt}{0pt}
        \multirow{17}{*}{\textbf{\shortstack{Melting\\Point}}} & Avalon & 0.9197 $\pm$ 0.0020 & 0.9742 $\pm$ 0.0023 & 0.9214 $\pm$ 0.0021 & 0.5275 $\pm$ 0.0911 & 0.5703 $\pm$ 0.0654 & 0.6519 $\pm$ 0.0476 \\
        \cmidrule(lr){2-8}
        & ECFP6 & 0.8570 $\pm$ 0.0054 & 0.8776 $\pm$ 0.0081 & 0.7510 $\pm$ 0.0159 & 0.3936 $\pm$ 0.0844 & 0.4762 $\pm$ 0.0823 & 0.5386 $\pm$ 0.0621 \\
        \cmidrule(lr){2-8}
        & Extended & 0.9111 $\pm$ 0.0035 & 0.9567 $\pm$ 0.0017 & 0.8561 $\pm$ 0.0221 & 0.4905 $\pm$ 0.0727 & 0.5520 $\pm$ 0.0704 & 0.6223 $\pm$ 0.0634 \\
        \cmidrule(lr){2-8}
        & FCFP4 & 0.7649 $\pm$ 0.0264 & 0.6908 $\pm$ 0.0218 & 0.6637 $\pm$ 0.0194 & 0.4686 $\pm$ 0.0818 & 0.4543 $\pm$ 0.0833 & 0.5294 $\pm$ 0.0639 \\
        \cmidrule(lr){2-8}
        & MACCS & 0.9166 $\pm$ 0.0068 & 0.9140 $\pm$ 0.0068 & 0.8400 $\pm$ 0.0137 & 0.6824 $\pm$ 0.0561 & 0.6926 $\pm$ 0.0543 & 0.6968 $\pm$ 0.0522 \\
        \cmidrule(lr){2-8}
        & Morgan & 0.8946 $\pm$ 0.0056 & 0.8226 $\pm$ 0.0048 & 0.7614 $\pm$ 0.0152 & 0.3515 $\pm$ 0.1464 & 0.4437 $\pm$ 0.0901 & 0.5555 $\pm$ 0.0628 \\
        \cmidrule(lr){2-8}
        & PCFP & 0.9052 $\pm$ 0.0054 & 0.9063 $\pm$ 0.0035 & 0.8063 $\pm$ 0.0254 & 0.5272 $\pm$ 0.0767 & 0.5713 $\pm$ 0.0678 & 0.6255 $\pm$ 0.0680 \\
        \cmidrule(lr){2-8}
        & rDesc & 0.9577 $\pm$ 0.0036 & 1.0000 $\pm$ 0.0000 & 0.9405 $\pm$ 0.0069 & 0.6751 $\pm$ 0.1127 & 0.6940 $\pm$ 0.1176 & 0.7033 $\pm$ 0.0610 \\
        \cmidrule(lr){2-8}
        & rPair & 0.7505 $\pm$ 0.0375 & 0.7734 $\pm$ 0.0423 & 0.7329 $\pm$ 0.0380 & 0.5578 $\pm$ 0.0907 & 0.5584 $\pm$ 0.0957 & 0.5763 $\pm$ 0.0908 \\
        \cmidrule(lr){2-8}
        & rTorsion & 0.5518 $\pm$ 0.0299 & 0.6199 $\pm$ 0.0360 & 0.5379 $\pm$ 0.0288 & 0.3594 $\pm$ 0.0667 & 0.3787 $\pm$ 0.0690 & 0.4442 $\pm$ 0.0759 \\
        \cmidrule(lr){2-8}
        & Standard & 0.7318 $\pm$ 0.0214 & 0.8575 $\pm$ 0.0270 & 0.8153 $\pm$ 0.0295 & 0.4800 $\pm$ 0.0468 & 0.5317 $\pm$ 0.0790 & 0.5877 $\pm$ 0.0807 \\
        \cmidrule(lr){2-8}
        & MACCS+Avalon & 0.9526 $\pm$ 0.0030 & 0.9912 $\pm$ 0.0007 & 0.9091 $\pm$ 0.0038 & 0.6672 $\pm$ 0.0607 & 0.6856 $\pm$ 0.0667 & 0.6983 $\pm$ 0.0555 \\
        \cmidrule(lr){2-8}
        & MACCS+ECFP6 & 0.9540 $\pm$ 0.0027 & 0.9759 $\pm$ 0.0006 & 0.8852 $\pm$ 0.0064 & 0.6867 $\pm$ 0.0544 & 0.7052 $\pm$ 0.0368 & 0.7130 $\pm$ 0.0477 \\
        \cmidrule(lr){2-8}
        & MACCS+Extended & 0.9557 $\pm$ 0.0021 & 0.9652 $\pm$ 0.0009 & 0.9273 $\pm$ 0.0087 & 0.6974 $\pm$ 0.0292 & 0.7080 $\pm$ 0.0163 & 0.7190 $\pm$ 0.0515 \\
        \cmidrule(lr){2-8}
        & MACCS+FCFP4 & 0.9375 $\pm$ 0.0069 & 0.9220 $\pm$ 0.0073 & 0.8682 $\pm$ 0.0152 & 0.7021 $\pm$ 0.0536 & 0.7090 $\pm$ 0.0521 & 0.7115 $\pm$ 0.0571 \\
        \cmidrule(lr){2-8}
        & MACCS+PCFP & 0.9533 $\pm$ 0.0028 & 0.9645 $\pm$ 0.0022 & 0.8900 $\pm$ 0.0145 & 0.7004 $\pm$ 0.0273 & 0.7337 $\pm$ 0.0363 & 0.7289 $\pm$ 0.0557 \\
        \cmidrule(lr){2-8}
        & MACCS+Standard & 0.9491 $\pm$ 0.0045 & 0.9617 $\pm$ 0.0035 & 0.8281 $\pm$ 0.0084 & 0.6871 $\pm$ 0.0471 & 0.7113 $\pm$ 0.0387 & 0.6857 $\pm$ 0.0494 \\
        \bottomrule
    \end{tabular}
\end{sidewaystable}

\newpage
\begin{sidewaystable}[ht!]
    \centering
    \caption{Mean Absolute Error from 5-fold Cross-Validation Sets for QSPR Models in Property Optimization Tasks (continued)}
    \vspace{5pt}
    \label{tab:table_s4}
    \renewcommand{\arraystretch}{1.3}  
    \begin{tabular}{@{}cccccccc@{}}
        \toprule
        \multirow{2}{*}{\textbf{Task}} & \multirow{2}{*}{\textbf{Descriptor}} 
          & \multicolumn{3}{c}{\textbf{Train MAE}} 
          & \multicolumn{3}{c}{\textbf{Test MAE}} \\
          \cmidrule(lr){3-8}
         & & \textbf{RF} & \textbf{XGB} & \textbf{LGBM} & \textbf{RF} & \textbf{XGB} & \textbf{LGBM} \\
        \specialrule{1.5pt}{0pt}{0pt}
        
        \multirow{17}{*}{\textbf{Viscosity}} & Avalon & 0.0522 $\pm$ 0.0008 & 0.0492 $\pm$ 0.0010 & 0.0626 $\pm$ 0.0014 & 0.1144 $\pm$ 0.0073 & 0.1202 $\pm$ 0.0062 & 0.1141 $\pm$ 0.0055 \\
        \cmidrule(lr){2-8}
        & ECFP6 & 0.0493 $\pm$ 0.0005 & 0.0264 $\pm$ 0.0006 & 0.0304 $\pm$ 0.0005 & 0.1100 $\pm$ 0.0056 & 0.0946 $\pm$ 0.0075 & 0.0937 $\pm$ 0.0062 \\
        \cmidrule(lr){2-8}
        & Extended & 0.0520 $\pm$ 0.0009 & 0.0391 $\pm$ 0.0011 & 0.0575 $\pm$ 0.0010 & 0.1128 $\pm$ 0.0049 & 0.1068 $\pm$ 0.0054 & 0.1032 $\pm$ 0.0049 \\
        \cmidrule(lr){2-8}
        & FCFP4 & 0.0726 $\pm$ 0.0011 & 0.0585 $\pm$ 0.0016 & 0.1011 $\pm$ 0.0021 & 0.1532 $\pm$ 0.0088 & 0.1468 $\pm$ 0.0095 & 0.1593 $\pm$ 0.0076 \\
        \cmidrule(lr){2-8}
        & MACCS & 0.0531 $\pm$ 0.0010 & 0.0440 $\pm$ 0.0010 & 0.0608 $\pm$ 0.0006 & 0.1105 $\pm$ 0.0074 & 0.1081 $\pm$ 0.0075 & 0.1061 $\pm$ 0.0067 \\
        \cmidrule(lr){2-8}
        & Morgan & 0.0503 $\pm$ 0.0006 & 0.0186 $\pm$ 0.0002 & 0.0420 $\pm$ 0.0013 & 0.1104 $\pm$ 0.0043 & 0.0913 $\pm$ 0.0071 & 0.1009 $\pm$ 0.0050 \\
        \cmidrule(lr){2-8}
        & PCFP & 0.0390 $\pm$ 0.0009 & 0.0202 $\pm$ 0.0007 & 0.0320 $\pm$ 0.0007 & 0.0930 $\pm$ 0.0050 & 0.0852 $\pm$ 0.0051 & 0.0805 $\pm$ 0.0045 \\
        \cmidrule(lr){2-8}
        & rDesc & 0.0247 $\pm$ 0.0002 & 0.0028 $\pm$ 0.0001 & 0.0105 $\pm$ 0.0006 & 0.0640 $\pm$ 0.0034 & 0.0594 $\pm$ 0.0031 & 0.0468 $\pm$ 0.0023 \\
        \cmidrule(lr){2-8}
        & rPair & 0.0436 $\pm$ 0.0009 & 0.0037 $\pm$ 0.0001 & 0.0165 $\pm$ 0.0004 & 0.0902 $\pm$ 0.0030 & 0.0688 $\pm$ 0.0046 & 0.0722 $\pm$ 0.0060 \\
        \cmidrule(lr){2-8}
        & rTorsion & 0.1239 $\pm$ 0.0017 & 0.0681 $\pm$ 0.0019 & 0.1313 $\pm$ 0.0008 & 0.1782 $\pm$ 0.0082 & 0.1364 $\pm$ 0.0069 & 0.1885 $\pm$ 0.0053 \\
        \cmidrule(lr){2-8}
        & Standard & 0.0619 $\pm$ 0.0010 & 0.0490 $\pm$ 0.0003 & 0.0701 $\pm$ 0.0009 & 0.1239 $\pm$ 0.0045      & 0.1168 $\pm$ 0.0037 & 0.1155 $\pm$ 0.0035 \\
        \cmidrule(lr){2-8}
        & MACCS+Avalon & 0.0437 $\pm$ 0.0006 & 0.0350 $\pm$ 0.0005 & 0.0367 $\pm$ 0.0006 & 0.0972 $\pm$ 0.0065 & 0.0985 $\pm$ 0.0070 & 0.0916 $\pm$ 0.0047 \\
        \cmidrule(lr){2-8}
        & MACCS+ECFP6 & 0.0368 $\pm$ 0.0003 & 0.0175 $\pm$ 0.0007 & 0.0271 $\pm$ 0.0005 & 0.0874 $\pm$ 0.0047 & 0.0820 $\pm$ 0.0027 & 0.0768 $\pm$ 0.0044 \\
        \cmidrule(lr){2-8}
        & MACCS+Extended & 0.0424 $\pm$ 0.0008 & 0.0214 $\pm$ 0.0006 & 0.0421 $\pm$ 0.0007 & 0.0982 $\pm$ 0.0046 & 0.0906 $\pm$ 0.0045 & 0.0854 $\pm$ 0.0032 \\
        \cmidrule(lr){2-8}
        & MACCS+FCFP4 & 0.0518 $\pm$ 0.0010 & 0.0387 $\pm$ 0.0012 & 0.0548 $\pm$ 0.0006 & 0.1069 $\pm$ 0.0075 & 0.1026 $\pm$ 0.0069 & 0.1007 $\pm$ 0.0064 \\
        \cmidrule(lr){2-8}
        & MACCS+PCFP & 0.0362 $\pm$ 0.0008 & 0.0161 $\pm$ 0.0005 & 0.0257 $\pm$ 0.0004 & 0.0869 $\pm$ 0.0064 & 0.0797 $\pm$ 0.0044 & 0.0735 $\pm$ 0.0044 \\
        \cmidrule(lr){2-8}
        & MACCS+Standard & 0.0415 $\pm$ 0.0008 & 0.0219 $\pm$ 0.0003 & 0.0327 $\pm$ 0.0008 & 0.0957 $\pm$ 0.0050 & 0.0897 $\pm$ 0.0059 & 0.0861 $\pm$ 0.0033 \\
        \bottomrule
    \end{tabular}
\end{sidewaystable}

\newpage
\begin{sidewaystable}[ht!]
    \centering
    Table S4: Mean Absolute Error from 5-fold Cross-Validation Sets for QSPR Models in Property Optimization Tasks (continued)
    \vspace{5pt}
    \label{tab:table_s4_2}
    \renewcommand{\arraystretch}{1.3}  
    \begin{tabular}{@{}cccccccc@{}}
        \toprule
        \multirow{2}{*}{\textbf{Task}} & \multirow{2}{*}{\textbf{Descriptor}} 
          & \multicolumn{3}{c}{\textbf{Train MAE}} 
          & \multicolumn{3}{c}{\textbf{Test MAE}} \\
          \cmidrule(lr){3-8}
         & & \textbf{RF} & \textbf{XGB} & \textbf{LGBM} & \textbf{RF} & \textbf{XGB} & \textbf{LGBM} \\
        \specialrule{1.5pt}{0pt}{0pt}
        
        \multirow{17}{*}{\textbf{\shortstack{Vapor\\Pressure}}} & Avalon & 0.4692 $\pm$ 0.0088 & 0.1985 $\pm$ 0.0102 & 0.4304 $\pm$ 0.0100 & 1.0501 $\pm$ 0.0681 & 1.0450 $\pm$ 0.0520 & 1.0088 $\pm$ 0.0602 \\
		\cmidrule(lr){2-8}
		& ECFP6 & 0.5651 $\pm$ 0.0148 & 0.1775 $\pm$ 0.0063 & 0.5391 $\pm$ 0.0094 & 1.2634 $\pm$ 0.0440 & 1.2401 $\pm$ 0.0309 & 1.1910 $\pm$ 0.0494 \\
		\cmidrule(lr){2-8}
		& Extended & 0.4584 $\pm$ 0.0125 & 0.1496 $\pm$ 0.0067 & 0.4280 $\pm$ 0.0125 & 0.9944 $\pm$ 0.0756 & 0.9947 $\pm$ 0.0637 & 0.9589 $\pm$ 0.0491 \\
		\cmidrule(lr){2-8}
		& FCFP4 & 0.6517 $\pm$ 0.0175 & 0.6963 $\pm$ 0.0231 & 0.7441 $\pm$ 0.0164 & 1.3300 $\pm$ 0.0905 & 1.3173 $\pm$ 0.0714 & 1.3031 $\pm$ 0.0683 \\
		\cmidrule(lr){2-8}
		& MACCS & 0.5286 $\pm$ 0.0101 & 0.3010 $\pm$ 0.0099 & 0.7654 $\pm$ 0.0103 & 1.0599 $\pm$ 0.0444 & 1.0588 $\pm$ 0.0349 & 1.0414 $\pm$ 0.0492 \\
		\cmidrule(lr){2-8}
		& Morgan & 0.5679 $\pm$ 0.0162 & 0.1401 $\pm$ 0.0062 & 0.6175 $\pm$ 0.0149 & 1.2547 $\pm$ 0.0482 & 1.2341 $\pm$ 0.0307 & 1.1802 $\pm$ 0.0463 \\
		\cmidrule(lr){2-8}
		& PCFP & 0.3609 $\pm$ 0.0064 & 0.1956 $\pm$ 0.0072 & 0.5857 $\pm$ 0.0158 & 0.8490 $\pm$ 0.0542 & 0.8208 $\pm$ 0.0411 & 0.8344 $\pm$ 0.0405 \\
		\cmidrule(lr){2-8}
		& rDesc & 0.2520 $\pm$ 0.0038 & 0.0041 $\pm$ 0.0010 & 0.1438 $\pm$ 0.0059 & 0.6859 $\pm$ 0.0466 & 0.6833 $\pm$ 0.0548 & 0.6332 $\pm$ 0.0386 \\
		\cmidrule(lr){2-8}
		& rPair & 0.4027 $\pm$ 0.0088 & 0.0298 $\pm$ 0.0024 & 0.3212 $\pm$ 0.0106 & 0.9009 $\pm$ 0.0495 & 0.8180 $\pm$ 0.0343 & 0.8178 $\pm$ 0.0448 \\
		\cmidrule(lr){2-8}
		& rTorsion & 0.8364 $\pm$ 0.0126 & 0.2559 $\pm$ 0.0103 & 0.9203 $\pm$ 0.0196 & 1.3625 $\pm$ 0.0537 & 1.1543 $\pm$ 0.0569 & 1.2874 $\pm$ 0.0614 \\
		\cmidrule(lr){2-8}
		& Standard & 0.5128 $\pm$ 0.0129 & 0.1897 $\pm$ 0.0118 & 0.5500 $\pm$ 0.0156 & 1.0766 $\pm$ 0.0614 & 1.0628 $\pm$ 0.0680 & 1.0344 $\pm$ 0.0611 \\
		\cmidrule(lr){2-8}
		& MACCS+Avalon & 0.4119 $\pm$ 0.0090 & 0.1430 $\pm$ 0.0101 & 0.4192 $\pm$ 0.0107 & 0.9417 $\pm$ 0.0617 & 0.9327 $\pm$ 0.0497 & 0.8920 $\pm$ 0.0537 \\
		\cmidrule(lr){2-8}
		& MACCS+ECFP6 & 0.4066 $\pm$ 0.0078 & 0.0780 $\pm$ 0.0064 & 0.4506 $\pm$ 0.0058 & 0.9774 $\pm$ 0.0413 & 0.9410 $\pm$ 0.0413 & 0.9105 $\pm$ 0.0471 \\
		\cmidrule(lr){2-8}
		& MACCS+Extended & 0.4005 $\pm$ 0.0086 & 0.1132 $\pm$ 0.0068 & 0.3525 $\pm$ 0.0084 & 0.9269 $\pm$ 0.0361 & 0.8974 $\pm$ 0.0361 & 0.8536 $\pm$ 0.0382 \\
		\cmidrule(lr){2-8}
		& MACCS+FCFP4 & 0.4408 $\pm$ 0.0083 & 0.1565 $\pm$ 0.0074 & 0.5575 $\pm$ 0.0119 & 1.0008 $\pm$ 0.0472 & 0.9832 $\pm$ 0.0469 & 0.9600 $\pm$ 0.0493 \\
		\cmidrule(lr){2-8}
		& MACCS+PCFP & 0.3379 $\pm$ 0.0043 & 0.1405 $\pm$ 0.0036 & 0.5140 $\pm$ 0.0115 & 0.8278 $\pm$ 0.0435 & 0.8046 $\pm$ 0.0330 & 0.7909 $\pm$ 0.0354 \\
		\cmidrule(lr){2-8}
		& MACCS+Standard & 0.4231 $\pm$ 0.0079 & 0.1770 $\pm$ 0.0078 & 0.4443 $\pm$ 0.0072 & 0.9564 $\pm$ 0.0318 & 0.9289 $\pm$ 0.0346 & 0.9045 $\pm$ 0.0474 \\

        \bottomrule
    \end{tabular}
\end{sidewaystable}

\newpage
\begin{sidewaystable}[ht!]
    \centering
    Table S4: Mean Absolute Error from 5-fold Cross-Validation Sets for QSPR Models in Property Optimization Tasks (continued)
    \vspace{5pt}
    \label{tab:table_s4_3}
    \renewcommand{\arraystretch}{1.3}  
    \begin{tabular}{@{}cccccccc@{}}
        \toprule
        \multirow{2}{*}{\textbf{Task}} & \multirow{2}{*}{\textbf{Descriptor}} 
          & \multicolumn{3}{c}{\textbf{Train MAE}} 
          & \multicolumn{3}{c}{\textbf{Test MAE}} \\
          \cmidrule(lr){3-8}
         & & \textbf{RF} & \textbf{XGB} & \textbf{LGBM} & \textbf{RF} & \textbf{XGB} & \textbf{LGBM} \\
        \specialrule{1.5pt}{0pt}{0pt}
        
        \multirow{17}{*}{\textbf{\shortstack{Boiling\\Point}}} & Avalon & 17.3877 $\pm$ 0.1726 & 20.0135 $\pm$ 0.1161 & 26.0231 $\pm$ 0.3516 & 32.8276 $\pm$ 1.1045 & 34.1708 $\pm$ 1.0398 & 35.1695 $\pm$ 0.6493 \\
		\cmidrule(lr){2-8}
		& ECFP6 & 24.4987 $\pm$ 0.2474 & 30.5057 $\pm$ 0.4642 & 30.0908 $\pm$ 0.6461 & 42.0744 $\pm$ 0.9848 & 42.6827 $\pm$ 1.4124 & 39.5387 $\pm$ 0.6686 \\
		\cmidrule(lr){2-8}
		& Extended & 23.0221 $\pm$ 0.2164 & 21.6051 $\pm$ 0.1174 & 27.8535 $\pm$ 0.4074 & 34.2761 $\pm$ 1.1223 & 32.6260 $\pm$ 1.1903 & 33.5599 $\pm$ 0.7087 \\
		\cmidrule(lr){2-8}
		& FCFP4 & 32.0229 $\pm$ 0.3083 & 23.9116 $\pm$ 0.3165 & 30.5974 $\pm$ 0.3569 & 45.5891 $\pm$ 1.1837 & 42.0005 $\pm$ 1.2836 & 43.0536 $\pm$ 1.2299 \\
		\cmidrule(lr){2-8}
		& MACCS & 21.4766 $\pm$ 0.1654 & 17.4496 $\pm$ 0.1401 & 25.6455 $\pm$ 0.3146 & 34.2512 $\pm$ 0.5930 & 32.9908 $\pm$ 0.6604 & 34.4723 $\pm$ 0.6426 \\
		\cmidrule(lr){2-8}
		& Morgan & 26.3002 $\pm$ 0.2665 & 31.9553 $\pm$ 0.3493 & 28.9075 $\pm$ 0.5342 & 42.3722 $\pm$ 1.1009 & 42.2758 $\pm$ 1.1023 & 38.3494 $\pm$ 0.6076 \\
		\cmidrule(lr){2-8}
		& PCFP & 20.3719 $\pm$ 0.5296 & 15.5003 $\pm$ 0.3272 & 18.3211 $\pm$ 0.3023 & 29.8851 $\pm$ 0.6985 & 26.4832 $\pm$ 0.9790 & 24.3217 $\pm$ 0.8834 \\
		\cmidrule(lr){2-8}
		& rDesc & 6.6924 $\pm$ 0.0830 & 4.1586 $\pm$ 0.1406 & 8.0413 $\pm$ 0.2153 & 16.3748 $\pm$ 0.5560 & 14.9647 $\pm$ 0.5258 & 14.8977 $\pm$ 0.5842 \\
		\cmidrule(lr){2-8}
		& rPair & 15.6009 $\pm$ 0.5103 & 6.0165 $\pm$ 0.4301 & 13.6085 $\pm$ 0.5235 & 26.1700 $\pm$ 0.7893 & 22.7767 $\pm$ 1.0517 & 20.9254 $\pm$ 0.9880 \\
		\cmidrule(lr){2-8}
		& rTorsion & 32.3111 $\pm$ 0.4193 & 11.4244 $\pm$ 0.3928 & 19.6784 $\pm$ 0.4212 & 42.0223 $\pm$ 0.9321 & 27.3570 $\pm$ 1.1061 & 29.5150 $\pm$ 0.9795 \\
		\cmidrule(lr){2-8}
		& Standard & 28.1876 $\pm$ 0.2905 & 20.8710 $\pm$ 0.2774 & 24.7708 $\pm$ 0.3872 & 38.7724 $\pm$ 0.8570 & 33.6532 $\pm$ 0.9018 & 33.4765 $\pm$ 0.6688 \\
		\cmidrule(lr){2-8}
		& MACCS+Avalon & 14.3461 $\pm$ 0.0902 & 8.6000 $\pm$ 0.0635 & 23.8060 $\pm$ 0.2380 & 27.6977 $\pm$ 0.5893 & 25.4484 $\pm$ 0.8306 & 29.3114 $\pm$ 0.6240 \\
		\cmidrule(lr){2-8}
		& MACCS+ECFP6 & 14.6715 $\pm$ 0.0718 & 5.3615 $\pm$ 0.0365 & 13.3142 $\pm$ 0.1921 & 29.1097 $\pm$ 0.7841 & 24.5718 $\pm$ 0.5995 & 27.5992 $\pm$ 0.7809 \\
		\cmidrule(lr){2-8}
		& MACCS+Extended & 14.2221 $\pm$ 0.0996 & 9.1373 $\pm$ 0.0861 & 20.7150 $\pm$ 0.2979 & 26.9891 $\pm$ 0.7866 & 25.4066 $\pm$ 0.8132 & 26.7427 $\pm$ 0.6741 \\
		\cmidrule(lr){2-8}
		& MACCS+FCFP4 & 17.2982 $\pm$ 0.1390 & 12.1720 $\pm$ 0.1412 & 21.6963 $\pm$ 0.2631 & 30.5750 $\pm$ 0.6446 & 29.1147 $\pm$ 0.6753 & 29.2181 $\pm$ 0.7360 \\
		\cmidrule(lr){2-8}
		& MACCS+PCFP & 16.8144 $\pm$ 0.3174 & 12.4125 $\pm$ 0.4001 & 13.1975 $\pm$ 0.2459 & 26.5555 $\pm$ 0.9004 & 22.0956 $\pm$ 0.7377 & 21.3474 $\pm$ 0.8277 \\
		\cmidrule(lr){2-8}
		& MACCS+Standard & 15.0954 $\pm$ 0.1055 & 9.0461 $\pm$ 0.1260 & 18.4586 $\pm$ 0.1537 & 28.4801 $\pm$ 1.0979 & 25.4917 $\pm$ 0.7915 & 26.9618 $\pm$ 0.6092 \\
        \bottomrule
    \end{tabular}
\end{sidewaystable}

\newpage
\begin{sidewaystable}[ht!]
    \centering
    Table S4: Mean Absolute Error from 5-fold Cross-Validation Sets for QSPR Models in Property Optimization Tasks (continued)
    \vspace{5pt}
    \label{tab:table_s4_4}
    \renewcommand{\arraystretch}{1.3}  
    \begin{tabular}{@{}cccccccc@{}}
        \toprule
        \multirow{2}{*}{\textbf{Task}} & \multirow{2}{*}{\textbf{Descriptor}} 
          & \multicolumn{3}{c}{\textbf{Train MAE}} 
          & \multicolumn{3}{c}{\textbf{Test MAE}} \\
          \cmidrule(lr){3-8}
         & & \textbf{RF} & \textbf{XGB} & \textbf{LGBM} & \textbf{RF} & \textbf{XGB} & \textbf{LGBM} \\
        \specialrule{1.5pt}{0pt}{0pt}
        
        \multirow{17}{*}{\textbf{\shortstack{Melting\\Point}}} & Avalon & 18.9529 $\pm$ 0.2190 & 11.0297 $\pm$ 0.3442 & 21.6631 $\pm$ 0.2782 & 42.0073 $\pm$ 1.2201 & 40.7917 $\pm$ 0.9321 & 38.5489 $\pm$ 0.6548 \\
		\cmidrule(lr){2-8}
		& ECFP6 & 27.5477 $\pm$ 0.3684 & 24.4285 $\pm$ 0.4235 & 32.0360 $\pm$ 0.5225 & 48.2928 $\pm$ 1.2862 & 46.2857 $\pm$ 1.2494 & 45.7972 $\pm$ 1.2585 \\
		\cmidrule(lr){2-8}
		& Extended & 19.8196 $\pm$ 0.2086 & 15.6583 $\pm$ 0.3000 & 21.8304 $\pm$ 0.5045 & 41.5633 $\pm$ 1.2805 & 40.2498 $\pm$ 1.6281 & 38.8988 $\pm$ 1.3600 \\
		\cmidrule(lr){2-8}
		& FCFP4 & 25.3407 $\pm$ 0.2453 & 32.5568 $\pm$ 0.3021 & 35.9154 $\pm$ 0.3061 & 47.1176 $\pm$ 1.4401 & 47.3897 $\pm$ 1.3207 & 45.6431 $\pm$ 1.3358 \\
		\cmidrule(lr){2-8}
		& MACCS & 19.4993 $\pm$ 0.2957 & 20.9204 $\pm$ 0.7916 & 27.1396 $\pm$ 0.2325 & 39.5261 $\pm$ 1.0891 & 38.6089 $\pm$ 0.9812 & 38.2426 $\pm$ 0.9642 \\
		\cmidrule(lr){2-8}
		& Morgan & 22.0873 $\pm$ 0.2921 & 31.8504 $\pm$ 0.3893 & 31.7288 $\pm$ 0.5407 & 46.6470 $\pm$ 1.3454 & 47.2710 $\pm$ 1.3560 & 44.5562 $\pm$ 0.9840 \\
		\cmidrule(lr){2-8}
		& PCFP & 20.5629 $\pm$ 0.5395 & 27.0725 $\pm$ 1.1841 & 23.6958 $\pm$ 0.2525 & 39.5804 $\pm$ 0.6510 & 39.7526 $\pm$ 0.9859 & 36.6846 $\pm$ 1.3655 \\
		\cmidrule(lr){2-8}
		& rDesc & 13.3164 $\pm$ 0.1626 & 0.0058 $\pm$ 0.0009 & 16.1765 $\pm$ 0.2090 & 34.9355 $\pm$ 1.4013 & 33.2154 $\pm$ 1.6414 & 32.8580 $\pm$ 0.8935 \\
		\cmidrule(lr){2-8}
		& rPair & 18.7145 $\pm$ 0.2305 & 13.4388 $\pm$ 0.5278 & 21.3272 $\pm$ 0.5309 & 40.0527 $\pm$ 0.9357 & 39.3359 $\pm$ 1.2217 & 38.1215 $\pm$ 1.0678 \\
		\cmidrule(lr){2-8}
		& rTorsion & 38.1780 $\pm$ 0.4031 & 32.6250 $\pm$ 0.2571 & 34.2105 $\pm$ 0.3135 & 53.8855 $\pm$ 1.1296 & 49.2769 $\pm$ 0.8152 & 46.6932 $\pm$ 0.9255 \\
		\cmidrule(lr){2-8}
		& Standard & 39.5215 $\pm$ 0.3954 & 23.7663 $\pm$ 0.1541 & 23.5328 $\pm$ 0.2843 & 48.8940 $\pm$ 0.5872 & 43.7250 $\pm$ 0.9898 & 41.2393 $\pm$ 1.1467 \\
		\cmidrule(lr){2-8}
		& MACCS+Avalon & 14.7519 $\pm$ 0.0882 & 6.5527 $\pm$ 0.4285 & 23.0656 $\pm$ 0.3470 & 37.4801 $\pm$ 0.7978 & 35.9206 $\pm$ 0.7576 & 35.5227 $\pm$ 0.7795 \\
		\cmidrule(lr){2-8}
		& MACCS+ECFP6 & 14.7570 $\pm$ 0.1204 & 13.2372 $\pm$ 0.4674 & 24.2530 $\pm$ 0.3495 & 37.2167 $\pm$ 0.9317 & 36.3039 $\pm$ 0.6946 & 35.9733 $\pm$ 0.7623 \\
		\cmidrule(lr){2-8}
		& MACCS+Extended & 14.5095 $\pm$ 0.1376 & 15.5373 $\pm$ 0.5970 & 17.7796 $\pm$ 0.3612 & 36.5525 $\pm$ 0.9622 & 35.7252 $\pm$ 0.7057 & 34.3025 $\pm$ 1.0332 \\
		\cmidrule(lr){2-8}
		& MACCS+FCFP4 & 15.4036 $\pm$ 0.2403 & 20.9573 $\pm$ 0.6580 & 22.8898 $\pm$ 0.2659 & 36.9120 $\pm$ 1.1305 & 37.0774 $\pm$ 0.7197 & 35.2945 $\pm$ 0.8894 \\
		\cmidrule(lr){2-8}
		& MACCS+PCFP & 13.9686 $\pm$ 0.1342 & 16.1028 $\pm$ 0.7215 & 19.3201 $\pm$ 0.2444 & 35.5762 $\pm$ 0.8548 & 34.3065 $\pm$ 0.8874 & 33.2599 $\pm$ 1.0821 \\
		\cmidrule(lr){2-8}
		& MACCS+Standard & 14.9254 $\pm$ 0.1279 & 15.1921 $\pm$ 0.4789 & 29.0459 $\pm$ 0.2215 & 37.3068 $\pm$ 0.8725 & 36.2012 $\pm$ 0.5919 & 36.9388 $\pm$ 0.9093 \\
        \bottomrule
    \end{tabular}
\end{sidewaystable}

\newpage
\begin{table}
    \centering
    \caption{Model Performance from 5-fold Cross-Validation Sets for Graph-based D-MPNN}
    \label{tab:table_s5}
    \renewcommand{\arraystretch}{1.3}
    \begin{tabular}{cccccc}
        \toprule
        \textbf{Model} & \textbf{Task} & \textbf{Train R-squared} & \textbf{Train MAE} & \textbf{Test R-squared} & \textbf{Test MAE} \\
        \specialrule{1.5pt}{0pt}{0pt}
        \multirow{4}{*}{Graph/D-MPNN} & Viscosity & 0.9070 $\pm$ 0.0199 & 0.1150 $\pm$ 0.0121 & 0.9609 $\pm$ 0.0187 & 0.0666 $\pm$ 0.0227 \\
		\cmidrule(lr){2-6}
		& Vapor pressure & 0.9891 $\pm$ 0.0033 & 0.2573 $\pm$ 0.0247 & 0.9952 $\pm$ 0.0008 & 0.2016 $\pm$ 0.0586 \\
		\cmidrule(lr){2-6}
		& Boiling Point & 0.9806 $\pm$ 0.0075 & 12.1769 $\pm$ 0.7928 & 0.9848 $\pm$ 0.0056 & 21.2934 $\pm$ 5.3825 \\
		\cmidrule(lr){2-6}
		& Melting Point & 0.9793 $\pm$ 0.0040 & 13.9387 $\pm$ 3.3526 & 0.9551 $\pm$ 0.0187 & 18.9523 $\pm$ 4.1407 \\
        \bottomrule
    \end{tabular}
\end{table}

\newpage
\begin{table}
    \centering
    \caption{Performance metrics of the best-found QSPR models for tasks}
    \label{tab:table_s6}
    \renewcommand{\arraystretch}{1.3}
    \begin{tabular}{cccccc}
        \toprule
        \textbf{Task} & \textbf{Model} & \textbf{Train R-squared} & \textbf{Train MAE} & \textbf{Test R-squared} & \textbf{Test MAE} \\
        \specialrule{1.5pt}{0pt}{0pt}
        Viscosity & rDesc/LGBM & 0.9985 $\pm$ 0.0002 & 0.0105 $\pm$ 0.0006 & 0.9605 $\pm$ 0.0060 & 0.0468 $\pm$ 0.0023 \\
		\cmidrule(lr){1-6}
		Vapor Pressure & Graph/D-MPNN & 0.9891 $\pm$ 0.0033 & 0.2573 $\pm$ 0.0247 & 0.9952 $\pm$ 0.0008 & 0.2016 $\pm$ 0.0586 \\
		\cmidrule(lr){1-6}
		Boiling Point & Graph/D-MPNN & 0.9806 $\pm$ 0.0075 & 12.1769 $\pm$ 0.7928 & 0.9848 $\pm$ 0.0056 & 21.2934 $\pm$ 5.3825 \\
		\cmidrule(lr){1-6}
		Melting Point & Graph/D-MPNN & 0.9793 $\pm$ 0.0040 & 13.9387 $\pm$ 3.3526 & 0.9551 $\pm$ 0.0187 & 18.9523 $\pm$ 4.1407 \\

        \bottomrule
    \end{tabular}
\end{table}

\newpage
\begin{sidewaystable}[ht!]
    \centering
    \caption{Predicted Properties of Amines Used in this work}
    \label{tab:table_s7}
    \vspace{5pt}
    \renewcommand{\arraystretch}{1.3}  
    \begin{tabular}{@{}ccccccccccc@{}}
        \toprule
        \textbf{\shortstack{Amine\\Type}} & \textbf{Name} & \textbf{\shortstack{Molecular\\Weight}} & \textbf{pKa} & \textbf{Viscosity} & \textbf{\shortstack{Vapor\\Pressure}} & \textbf{\shortstack{Boiling\\Point}} & \textbf{\shortstack{Melting\\Point}} & \textbf{RAscore} & \textbf{Price} & \textbf{\shortstack{Aqueous\\Solubility}} \\
        \specialrule{1.5pt}{0pt}{0pt}
        \multirow{6}{*}{Primary} & MEA & 61.084 & 9.160 & 0.728 & -0.438 & 132.005 & -3.635 & 0.991 & 3.035 & 1.690 \\
		& AHMPD & 121.136 & 7.637 & 1.185 & -6.003 & 326.387 & 125.803 & 0.987 & 4.887 & 1.010 \\
		& AMP & 89.138 & 8.817 & 0.643 & -1.777 & 154.140 & 19.075 & 0.981 & 4.323 & 1.267 \\
		& IPA & 59.112 & 2.814 & -0.429 & 2.734 & 45.598 & -93.031 & 0.987 & 4.004 & 1.692 \\
		& SAGE-01 & 130.231 & 11.793 & 0.718 & -3.901 & 207.633 & -50.750 & 0.953 & 4.680 & 0.333 \\
		& SAGE-02 & 103.163 & 10.538 & 0.142 & -0.920 & 221.044 & -3.753 & 0.991 & 2.723 & 0.706 \\
		\cmidrule(lr){1-11}
		\multirow{5}{*}{Secondary} & DEA & 105.137 & 9.096 & 1.261 & -2.993 & 253.185 & -14.309 & 0.994 & 2.515 & 1.247 \\
		& DA & 73.139 & 1.795 & -0.512 & 2.339 & 42.306 & -79.072 & 0.992 & 3.244 & 1.239 \\
		& IPAP & 117.192 & 9.936 & 0.748 & -1.402 & 186.172 & -19.294 & 0.993 & 3.077 & 0.693 \\
		& SAGE-03 & 100.162 & 11.912 & 0.521 & -2.778 & 166.675 & 15.283 & 0.868 & 4.449 & 0.961 \\
		& SAGE-04 & 160.260 & 11.000 & 0.386 & -3.240 & 260.011 & 25.658 & 0.984 & 4.312 & 0.905 \\
		\cmidrule(lr){1-11}
		\multirow{7}{*}{Tertiary} & MDEA & 119.164 & 7.969 & 0.940 & -2.764 & 237.189 & -15.973 & 0.995 & 4.244 & 0.985 \\
		& DEEA & 117.192 & 8.163 & 0.501 & -0.172 & 155.978 & -33.024 & 0.990 & 3.842 & 0.762 \\
		& 4DMA1B & 117.192 & 9.302 & 0.651 & 0.568 & 165.326 & -29.512 & 0.990 & 4.401 & 0.788 \\
		& 1DMA2P & 103.165 & 8.401 & 0.387 & 0.036 & 122.668 & -66.949 & 0.986 & 3.931 & 1.080 \\
		& DEAE-EO & 161.245 & 8.429 & 0.559 & -0.847 & 224.180 & -28.831 & 0.991 & 5.068 & 0.389 \\
		& SAGE-05 & 159.272 & 10.722 & 0.317 & -2.945 & 677.109 & 27.061 & 0.969 & 4.383 & 0.299 \\
		& SAGE-06 & 145.246 & 11.030 & 0.287 & -2.045 & 239.905 & -42.911 & 0.963 & 4.384 & 0.576 \\
		\cmidrule(lr){1-11}
		\multirow{11}{*}{Cyclic} & 2-MPZ & 100.165 & 10.178 & 0.435 & 0.474 & 128.757 & 62.466 & 0.980 & 5.769 & 1.085 \\
		& 2-PPE & 129.203 & 10.222 & 1.260 & -1.460 & 210.111 & 36.302 & 0.984 & 4.941 & 0.386 \\
		& HomoPZ & 100.165 & 10.995 & 0.516 & 1.185 & 88.906 & 128.928 & 0.945 & 6.405 & 1.115 \\
		& 1M-2PPE & 129.203 & 9.648 & 1.073 & -0.067 & 208.248 & 55.987 & 0.988 & 4.540 & 0.433 \\
		& EPZ & 114.192 & 9.508 & 0.258 & -0.011 & 156.735 & 4.751 & 0.990 & 3.977 & 0.907 \\
		& PZ & 86.138 & 10.461 & 0.470 & -0.179 & 96.078 & -5.808 & 0.975 & 5.796 & 1.181 \\
		& 1-MPZ & 100.165 & 9.502 & 0.219 & 0.542 & 123.059 & -1.550 & 0.984 & 4.968 & 0.924 \\
		& AEP & 129.207 & 9.622 & 0.485 & -1.270 & 221.794 & 71.807 & 0.990 & 4.236 & 0.838 \\
		& 1-(2HE)PP & 129.203 & 9.252 & 0.915 & -1.241 & 193.568 & 53.934 & 0.982 & 4.894 & 0.456 \\
		& SAGE-07 & 170.255 & 13.256 & 0.391 & -3.951 & 177.763 & 28.361 & 0.954 & 5.101 & -0.358 \\
		& SAGE-08 & 170.255 & 13.672 & 0.448 & -1.343 & 210.015 & 36.368 & 0.986 & 5.225 & -0.383 \\
		\cmidrule(lr){1-11}
		\multirow{4}{*}{Poly} & DETA & 103.169 & 10.272 & 0.818 & -1.234 & 204.826 & 26.278 & 0.964 & 4.096 & 1.386 \\
		& TETA & 146.238 & 10.364 & 0.626 & -3.424 & 277.009 & 117.589 & 0.968 & 4.465 & 1.118 \\
		& SAGE-09 & 158.245 & 13.127 & 0.449 & -1.168 & 236.530 & 0.726 & 0.956 & 4.840 & 0.436 \\
		& SAGE-10 & 158.245 & 12.859 & 0.375 & -3.155 & 250.238 & 63.089 & 0.968 & 4.454 & 0.709 \\
        \bottomrule
    \end{tabular}
\end{sidewaystable}

\newpage
\begin{sidewaystable}[ht!]
    \centering
    \caption{COSMO-RS Results of Amines Used in this work}
    \label{tab:table_s8}
    \vspace{5pt}
    \renewcommand{\arraystretch}{1.3}  
    \begin{tabular}{@{}ccccccccc@{}}
        \toprule
        \textbf{\shortstack{Amine\\Type}} & \textbf{Name} & \textbf{\shortstack{Molecular\\Weight}} & \textbf{\shortstack{CO\textsubscript{2} Henry's\\Law Constant}} & \textbf{\shortstack{CO\textsubscript{2}\\Solubility}} & \textbf{\shortstack{Flash\\Point}} & \textbf{\shortstack{Vaporization\\$\Delta$H}} & \textbf{\shortstack{Excess\\$\Delta$H}} & \textbf{\shortstack{Excess\\$\Delta$G}} \\
        \specialrule{1.5pt}{0pt}{0pt}
        \multirow{6}{*}{Primary} & MEA & 61.084 & 345.911 & -3.045 & 128.346 & 51.581 & -0.989 & -0.319 \\
		& AHMPD & 121.136 & 440.125 & -3.150 & 234.885 & 52.148 & -0.351 & -0.048 \\
		& AMP & 89.138 & 312.544 & -3.002 & 137.628 & 51.082 & -0.573 & 0.047 \\
		& IPA & 59.112 & 222.534 & -2.854 & -19.444 & 48.847 & -0.875 & 0.216 \\
		& SAGE-01 & 130.231 & 279.746 & -2.955 & 88.115 & 50.255 & -0.680 & 0.130 \\
		& SAGE-02 & 103.163 & 254.988 & -2.915 & 48.431 & 50.032 & -0.827 & 0.098 \\
        \cmidrule(lr){1-9}
		\multirow{5}{*}{Secondary} & DEA & 105.137 & 359.695 & -3.063 & 216.388 & 52.541 & -0.998 & -0.272 \\
		& DA & 73.139 & 206.572 & -2.823 & -14.443 & 48.984 & -0.716 & 0.505 \\
		& IPAP & 117.192 & 270.021 & -2.940 & 118.481 & 50.529 & -0.274 & 0.374 \\
		& SAGE-03 & 100.162 & 284.207 & -2.961 & 88.682 & 50.365 & -0.597 & 0.138 \\
		& SAGE-04 & 160.260 & 278.798 & -2.955 & 145.767 & 51.109 & -1.109 & -0.108 \\
        \cmidrule(lr){1-9}
		\multirow{7}{*}{Tertiary} & MDEA & 119.164 & 329.834 & -3.025 & 165.051 & 51.214 & -0.386 & 0.187 \\
		& DEEA & 117.192 & 265.029 & -2.932 & 86.602 & 50.088 & -0.295 & 0.473 \\
		& 4DMA1B & 117.192 & 261.044 & -2.925 & 98.842 & 50.419 & -0.459 & 0.369 \\
		& 1DMA2P & 103.165 & 249.698 & -2.906 & 36.519 & 49.507 & -0.615 & 0.354 \\
		& DEAE-EO & 161.245 & 284.424 & -2.963 & 123.121 & 50.122 & -0.180 & 0.447 \\
		& SAGE-05 & 159.272 & 261.068 & -2.927 & 112.541 & 50.370 & -0.744 & 0.143 \\
		& SAGE-06 & 145.246 & 265.298 & -2.933 & 85.482 & 50.463 & -0.834 & 0.149 \\
        \cmidrule(lr){1-9}
		\multirow{11}{*}{Cyclic} & 2-MPZ & 100.165 & 273.125 & -2.944 & 64.691 & 50.489 & -1.156 & -0.049 \\
		& 2-PPE & 129.203 & 290.767 & -2.972 & 150.194 & 50.651 & -0.273 & 0.285 \\
		& HomoPZ & 100.165 & 277.481 & -2.951 & 69.746 & 50.433 & -1.096 & -0.082 \\
		& 1M-2PPE & 129.203 & 288.993 & -2.969 & 128.004 & 50.410 & -0.264 & 0.387 \\
		& EPZ & 114.192 & 260.808 & -2.925 & 63.932 & 50.302 & -1.010 & 0.129 \\
		& PZ & 86.138 & 280.012 & -2.954 & 55.932 & 50.442 & -1.238 & -0.161 \\
		& 1-MPZ & 100.165 & 264.517 & -2.930 & 50.912 & 50.285 & -1.151 & 0.065 \\
		& AEP & 129.207 & 296.656 & -2.981 & 127.301 & 51.431 & -1.518 & -0.424 \\
		& 1-(2HE)PP & 129.203 & 282.591 & -2.960 & 121.231 & 50.400 & -0.340 & 0.358 \\
		& SAGE-07 & 170.255 & 294.048 & -2.978 & 153.061 & 50.671 & -0.780 & -0.021 \\
		& SAGE-08 & 170.255 & 290.983 & -2.974 & 158.541 & 50.689 & -0.711 & -0.005 \\
        \cmidrule(lr){1-9}
		\multirow{4}{*}{Poly} & DETA & 103.169 & 297.854 & -2.981 & 126.789 & 51.966 & -1.867 & -0.788 \\
		& TETA & 146.238 & 293.344 & -2.976 & 155.761 & 51.878 & -1.544 & -0.533 \\
		& SAGE-09 & 158.245 & 286.817 & -2.967 & 178.189 & 50.958 & -0.604 & -0.016 \\
		& SAGE-10 & 158.245 & 291.602 & -2.974 & 162.888 & 50.964 & -0.764 & -0.017 \\

        \bottomrule
    \end{tabular}
\end{sidewaystable}

\end{document}